\documentclass{article}

\PassOptionsToPackage{numbers}{natbib}
\usepackage{iclr2027_conference,times}
\iclrfinalcopy

\usepackage[utf8]{inputenc}
\usepackage[T1]{fontenc}
\usepackage{graphicx}
\usepackage{booktabs}
\usepackage{amsmath,amssymb}
\usepackage{xcolor}
\usepackage{hyperref}
\usepackage[capitalise,noabbrev]{cleveref}
\usepackage{microtype}
\usepackage{multirow}
\usepackage{caption}
\usepackage{enumitem}
\setlist{nosep,leftmargin=*}
\graphicspath{{figs/}}

\newif\ifcleanversion
\cleanversionfalse

\colorlet{reddish}{red!70}

\title{Beyond Resolution: \\ Object-to-Image Ratio Mismatch in Instance Retrieval}

\author{%
  Boaz Meivar\thanks{Equal contribution.}\\
  Tel Aviv University\\
  \texttt{boazmeivar@mail.tau.ac.il}
  \And
  Ofir Kedem\footnotemark[1]\\
  Tel Aviv University\\
  \texttt{ofirk@mail.tau.ac.il}
  \And
  Amit Edenzon\footnotemark[1]\\
  Bar-Ilan University\\
  \texttt{amit.edenzon@live.biu.ac.il}
  \AND
  Gal Chechik\\
  Bar-Ilan University\\
  \texttt{gal.chechik@biu.ac.il}
  \And
  Shai Avidan\\
  Tel Aviv University\\
  \texttt{avidan@eng.tau.ac.il}
}

\begin{document}
\maketitle

\begin{abstract}
Visual instance retrieval often fails when the same object appears at
different apparent sizes in the query and gallery. We show that the dominant
cause is usually not resolution loss but \emph{object-to-image} (O2I) ratio
mismatch: the object occupies different fractions of the two images. On a
controlled benchmark of 3{,}021 Objaverse objects rendered at five camera
distances, more than $80\%$ of the cross-distance degradation is attributable
to O2I mismatch rather than resolution for 9 of 12 pretrained backbones;
multi-scale architectures cut the resolution-only effect to single digits yet
remain equally susceptible. The failure is also asymmetric: tight queries
retrieve more reliably against wide gallery images than the reverse. Guided by
this analysis, query-side scale augmentation and an OWLv2 crop reranker reach
state of the art on ILIAS~100M (\textbf{29.2} mAP@1000 before reranking,
\textbf{42.0} after) without training or modifying the precomputed gallery
index, and a LoRA fine-tune matches the query-side gains at a single forward
pass, showing that O2I robustness is learnable.
\end{abstract}

\begin{figure}[!h]
  \centering
  \includegraphics[width=\linewidth]{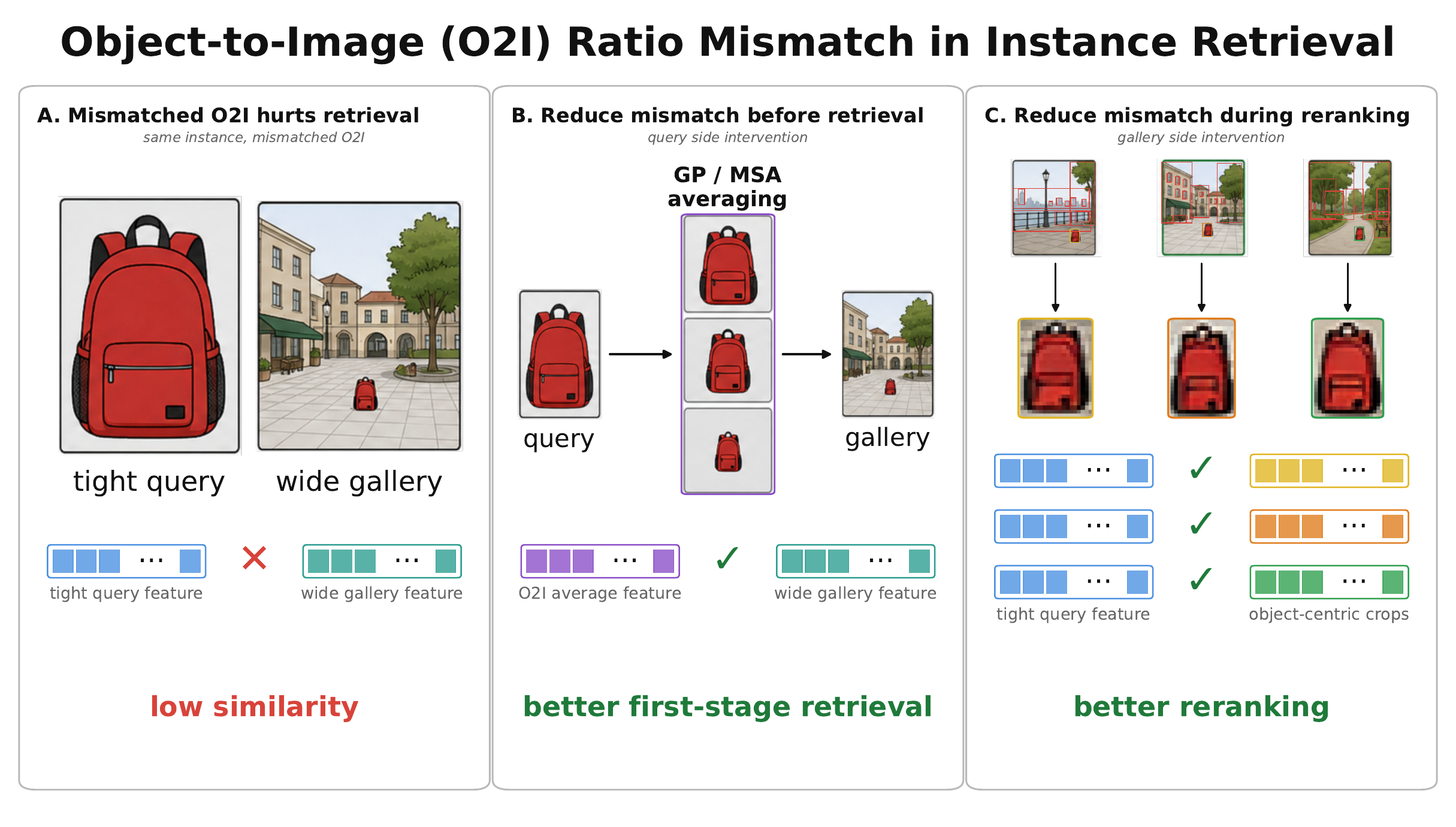}
  \caption{\textbf{We study how object-to-image (O2I) ratio mismatch
  degrades feature similarity (panel~A).} Motivated by the analysis
  results, we apply query-side (panel~B) and gallery-side
  (panel~C) interventions to reduce the O2I mismatch between query
  and gallery, and observe a significant improvement in retrieval
  outcome, reaching state of the art on the ILIAS~100M benchmark.}
  \label{fig:teaser}
\end{figure}


\section{Introduction}
\label{sec:intro}

Visual instance retrieval asks whether a gallery image contains the same
physical object as a query~\citep{ilias2025,radenovic2018revisiting,weyand2020GLDv2}.
A common modern pipeline encodes each image with a pretrained vision
backbone, normalizes the resulting global descriptor, and ranks gallery images
by some predefined metric like cosine similarity. This formulation is simple
and effective, but its failures are difficult to interpret: image pairs vary
simultaneously in viewpoint, illumination, occlusion, background, pose,
framing, and camera distance. Overall retrieval scores therefore reveal that a
model fails, but not which visual factor caused the failure.

A long-standing difficulty in image retrieval is the small target: an
object that occupies a small fraction of the image is hard to find, a
problem documented from classification~\citep{pawlowski2019needles} to
instance retrieval~\citep{green2025needle,choi2026patchify}. The failure
is commonly attributed to properties of the image itself: surrounding
clutter, limited object resolution, or lost detail. This view motivates
many standard responses to scale variation: multi-scale
features~\citep{he2015spatial,lin2017feature}, local
descriptors~\citep{tolias2016particular}, or architectures designed to
preserve spatial detail~\citep{liu2021swin,beyer2023flexivit,li2022mvitv2}.

We show that a major cause of this failure in image retrieval is the
\emph{mismatch} between the fractions the object occupies in the query
and in the gallery image.
The effect requires neither clutter nor resolution loss: with an identical
object view, a neutral background, and matched resolution
(\cref{sec:canvas}), the global feature vector itself changes considerably
when only the occupied fraction changes. \textbf{A small query object that
is matched by a small target retrieves reliably}, while an identical object
that is mismatched does not.

Two factors co-vary under scale change: when an image is zoomed out,
the object simultaneously becomes smaller relative to the frame (a lower
\emph{object-to-image} ratio, O2I~\citep{pawlowski2019needles}) and loses pixel density (lower effective
resolution). One might infer that targeting either factor is sufficient,
since the two are linearly coupled. However, we show in
\cref{sec:results} below that addressing resolution alone (e.g., with
multi-scale architectures) does not fix O2I-related retrieval failures.
Directly addressing O2I, without improving resolution, successfully
mitigates the failure mode (\cref{sec:augmentation}). We refer to
the dominant effect as \emph{object-to-image mismatch} (O2I mismatch). Unlike resolution loss,
object-to-image mismatch changes the image support from which the descriptor is
formed. A tightly framed object produces a descriptor whose input is mostly the
target object, whereas a wide image produces a descriptor in which the same
object is only a small part of the input. Thus, retrieval may fail not because
the object is visually unrecoverable, but because the same instance is compared
under incompatible image support. These two mechanisms are usually conflated
under ``scale mismatch'' but imply different interventions: resolution loss
calls for preserving or recovering object detail, whereas object-to-image
mismatch calls for comparing objects at compatible fractions of the input
image.

We test this distinction with a controlled synthetic benchmark of
3{,}021 Objaverse objects rendered at five camera distances with fixed
viewpoint and neutral background. Because only camera distance changes, we can
construct matched comparisons that isolate pixel resolution from
object-to-image mismatch. Across twelve backbones spanning self-supervised,
vision-language, supervised, multi-scale, and reconstruction-based training
paradigms, we find that the intuitive resolution explanation is not the
dominant factor. For 9 of 12 backbones tested, more than $80\%$ of the
cross-distance retrieval degradation is attributable to object-to-image mismatch
rather than resolution loss. Multi-scale architectures reduce the
resolution-only effect, but remain equally susceptible to object-to-image
mismatch: used as global descriptors, scale-specific architectures deliver
resolution invariance, but not invariance to the fraction of the input the
object occupies. We additionally observe that the failure is asymmetric: tight
queries retrieve more reliably against wide gallery images than wide queries
retrieve against tight gallery images.

Motivated by these findings, we apply simple interventions targeting
object-to-image mismatch on ILIAS~100M, including query-side scale augmentation
and an OWLv2 crop reranker. These methods compare objects at more compatible
fractions of the input image and reach state-of-the-art performance without
training a new backbone or modifying the precomputed gallery index. We also show that object-to-image
robustness is learnable: a LoRA fine-tune of DINOv3-L achieves similar gains to
query-side test-time interventions while requiring only a single forward pass
per image.

\noindent\textbf{Contributions.}
\begin{itemize}[leftmargin=*,itemsep=1pt,topsep=1pt]
  \item \textbf{We isolate the failure mode.} We construct a controlled benchmark (3{,}021 objects, five camera distances, fixed viewpoint) in which only the O2I ratio varies, and show that all 12 evaluated backbones degrade sharply under O2I mismatch, including architectures designed for multi-scale processing (Swin, FlexiViT, MViTv2).
  \item \textbf{We attribute it: resolution is generally not the bottleneck.} Equalizing the object-to-image ratio between query and gallery recovers more than $80\%$ of the cross-distance failure on 9 of 12 backbones tested; the residual is attributable to resolution loss, since all other factors are controlled by construction.
  \item \textbf{We identify its direction.} The failure is asymmetric: tight queries retrieve more reliably against wide gallery images than the reverse, revealing a directional property of how global descriptors encode object support.
  \item \textbf{We correct it at web scale, training-free.} Query-side scale augmentation reaches \textbf{29.2} mAP@1000 in first-stage retrieval and an OWLv2 crop reranker \textbf{42.0} after reranking, on ILIAS~100M, without training or modifying the precomputed gallery index.
  \item \textbf{We show it is learnable.} A LoRA fine-tune of DINOv3-L matches the test-time interventions at single-forward-pass cost, suggesting O2I invariance can be built into the representation by pretraining.
\end{itemize}

\section{Related Work}
\label{sec:related}

\paragraph{Instance retrieval benchmarks.}
Instance retrieval has traditionally been evaluated on landmark-centric
benchmarks such as Oxford and Paris
\citep{philbin2007object,philbin2008lost}, later revised to reduce label
noise and saturation \citep{radenovic2018revisiting}.
Google Landmarks~\citep{weyand2020GLDv2} scaled evaluation to
millions of images and broader visual diversity, while
ILIAS~\citep{ilias2025} targets instance-level retrieval at scale
with curated positives, hard distractors, and unsaturated
performance for modern foundation models.
These benchmarks are indispensable for measuring real-world retrieval, but
they entangle many nuisance factors at once, including viewpoint,
background, occlusion, and framing.
As a result, they are not well suited to answering a narrower scientific
question: how much of the failure comes specifically from
query--gallery O2I mismatch?

\paragraph{Controlled robustness and distribution shift.}
A broad literature studies robustness under distribution shift, but
typically through classification rather than retrieval.
ImageNet-C~\citep{hendrycks2019benchmarking} measures sensitivity
to corruptions such as blur, noise, and weather, while
ObjectNet~\citep{barbu2019objectnet} highlights the importance of
controlled evaluation under changes in viewpoint, background, and
pose.
Related work on shortcut learning argues that modern models often exploit
spurious or context-heavy cues that fail under shift
\citep{geirhos2020shortcut}.
In parallel, recent analyses of Vision Transformers show that
high-norm or artifact tokens can arise in low-information background
regions, indicating that global representations need not be purely
object-centric \citep{darcet2024registers}.
These results motivate studying apparent object size as a distinct factor:
when the object becomes small, the representation may devote a larger
fraction of its capacity to surrounding content.
However, this issue has rarely been isolated in instance retrieval.

\paragraph{Scale-aware representations.}
Multi-scale designs from spatial pyramid pooling and feature
pyramids~\citep{he2015spatial,lin2017feature} to
scale-equivariant
networks~\citep{worrall2019deepscalespaces,sosnovik2019scaleequivariant,ashiquzzaman2023scaleequivfourier}
and scale-aware transformers~\citep{lin2023scaleaware} treat
scale as a structural property to be modeled directly. They are
typically evaluated on classification or detection, not
image-to-image retrieval under controlled O2I mismatch.
Detection and localization have addressed scale explicitly for
decades, from scale-invariant interest
points~\citep{lowe2004distinctive} and image pyramids to
objectness scoring and open-vocabulary
detectors~\citep{minderer2024owlv2,liu2024grounding}. This
machinery relies on localized readout, which web-scale
first-stage retrieval cannot afford per gallery image; we
leverage it where it is affordable, at the reranking stage
(\cref{sec:rerank}).

\paragraph{Retrieval-time multi-scale inference and recent work.}
Regional/part-based
representations~\citep{tolias2016particular} and query
expansion/reranking~\citep{chum2007total,arandjelovic2012three}
operate on local structure or refine global rankings.
Recent work on patch-wise
retrieval~\citep{choi2026patchify} decomposes gallery images
into multi-scale grid patches and max-pools their similarities,
empirically finding that local matching helps most for small
and off-center objects (they also evaluate a
Grounding-DINO~\citep{liu2024grounding} region-proposal variant
as an upper-bound baseline). We identify the shared phenomenon these methods exploit: O2I
mismatch between query and gallery. Any intervention that
equalizes it yields consistent gains, including Patchify~\citep{choi2026patchify}, the
query-side scale augmentation of \cref{sec:augmentation}, and
the OWLv2 crop reranker of \cref{sec:rerank}.

\section{Controlled O2I Benchmark}
\label{sec:benchmark}

Real-world retrieval benchmarks
\citep{radenovic2018revisiting,ilias2025} vary O2I alongside
viewpoint, lighting, background, and partial occlusion, so a
backbone's failure on any given query cannot be attributed to
O2I specifically.
To measure O2I's isolated effect on the representation, we
construct a synthetic benchmark in which only the O2I ratio
varies and every other factor is held fixed by construction:
the same object, the same viewpoint, the same lighting, the
same (neutral) background, across a range of camera distances
(up to the mild perspective foreshortening that accompanies a
distance change under a perspective camera).

\subsection{Dataset Construction}

We render 3{,}021 objects from the Objaverse
dataset~\citep{deitke2023objaverse}, drawn from 120 categories.
Each object is rendered at five camera distances
$d \in \{1.5, 2.0, 2.5, 3.5, 5.0\}$ from a single canonical
isometric viewpoint (azimuth~$45^\circ$, elevation~$35.3^\circ$)
using a perspective camera with a $45^\circ$ vertical field of
view, producing 15{,}105 images at $512 \times 512$ resolution.
The reference distance $d{=}2.5$ is chosen so that a nominal object fills
roughly half the frame; the five distances span a $3.3{\times}$ range
in apparent object size. Each mesh is centered at the origin and
normalized to unit maximum bounding-box extent before rendering,
so the range of O2I ratios across the five distances is
identical for every object---a distance change corresponds to
the same apparent-size change regardless of the object's
original mesh scale.

Rendering uses GPU rasterization via
Kaolin~\citep{jatavallabhula2019kaolin} with three-point lighting
and a neutral gray background, ensuring that the dominant variation
across images of the same object is the fraction of the image it
occupies (residual perspective foreshortening aside).
This design isolates the O2I ratio from viewpoint, lighting, and
background---confounds that are unavoidable in real-world benchmarks.
Example renderings of one Objaverse object across the five
distances appear in \cref{app:benchmark_example}.

\subsection{Cross-Distance Retrieval Protocol}
\label{sec:protocol}

For each ordered pair of distances $(d_q, d_t)$ with $d_q \neq d_t$,
each of the 3{,}021 objects serves as a query (rendered at $d_q$).
The gallery is 15{,}101 items: the same object at $d_t$ plus all
other objects at all five distances. The query object's
renderings at the four remaining distances are masked to prevent
trivial self-matches.

This yields $N$ ranks per pair, from which we compute
R@1, R@5, R@10, and mean reciprocal rank (MRR). We aggregate
across the 20 off-diagonal pairs by the directional O2I ratio
$\mathrm{O2I}_q / \mathrm{O2I}_t = (d_t/d_q)^2$ (values ${>}1$
when the query is closer/larger than the target, ${<}1$ when
farther/smaller).

\subsection{Backbones Evaluated}
\label{sec:backbones}

We cover twelve backbones spanning five training paradigms:
self-supervised (DINOv2-L, DINOv2-G~\citep{oquab2024dinov2},
DINOv3-L~\citep{simeoni2025dinov3}), vision-language contrastive
(CLIP-L~\citep{radford2021learning},
SigLIP-2-L~\citep{tschannen2025siglip2}), supervised
classification (ConvNeXt V2-H, ConvNeXt V2-B~\citep{woo2023convnextv2},
ResNet-50~\citep{he2016deep}), hierarchical/multi-scale
architectures designed to handle scale variation
(Swin-L~\citep{liu2021swin},
FlexiViT-L~\citep{beyer2023flexivit},
MViTv2-L~\citep{li2022mvitv2}), and 3D reconstruction
(DUSt3R-L~\citep{wang2024dust3r}). Including the hierarchical
family lets us test whether architectures built for scale
robustness also confer robustness to O2I mismatch.
Implementation details are listed in \cref{app:backbones}.

\section{O2I Robustness Analysis}
\label{sec:results}

We now ask what the isolated effect of O2I mismatch is on
current retrieval backbones. We first show that every backbone
degrades with cross-distance mismatch, then disentangle the
contribution of O2I from that of pixel-resolution loss
(\cref{sec:disentangle}), and finally surface a directional
asymmetry of the failure (\cref{sec:asymmetry}).

\subsection{All Models Degrade Under O2I Mismatch}

\begin{figure}[t]
  \centering
  \includegraphics[width=\linewidth]{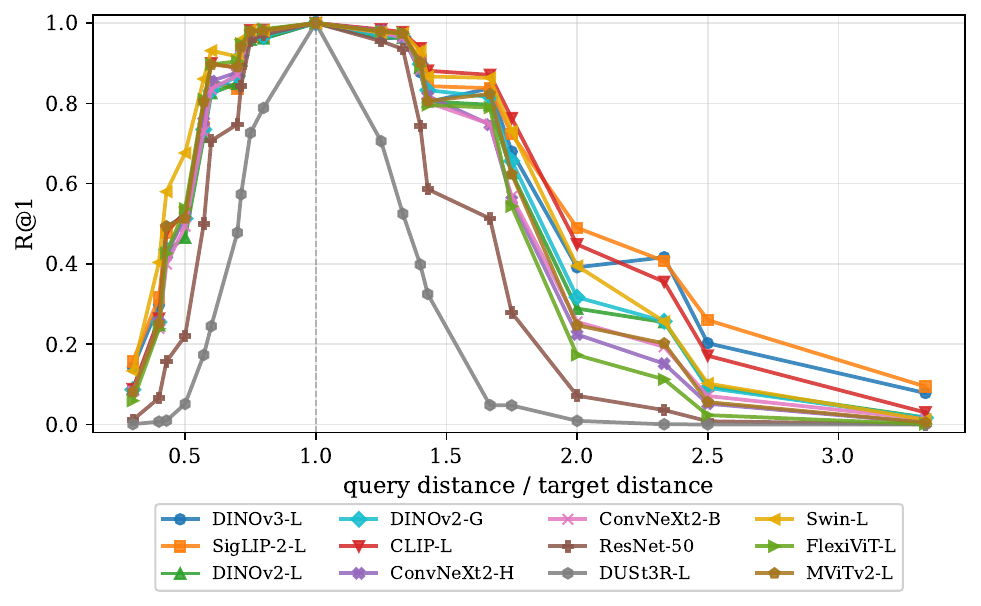}
  \caption{\textbf{A comparison of retrieval performance of
  twelve backbones as the O2I ratio changes.} The y-axis is
  cross-distance R@1; the x-axis is the camera-distance ratio
  $d_q/d_t$ (ratio $=1$ is self-retrieval). All backbones degrade
  with O2I mismatch, including those built for scale invariance
  (Swin-L, FlexiViT-L, MViTv2-L). Performance degrades for both
  directions, losing 0.5--0.9 R@1 at the extremes.}
  \label{fig:retrieval_vs_o2i}
\end{figure}

We first establish the cross-distance failure curve.
\cref{fig:retrieval_vs_o2i} shows that retrieval peaks at
self-distance and degrades as the camera-distance ratio diverges
in either direction. At the most extreme ratio
($3.3{\times}$ camera-distance change) even the strongest
backbone (SigLIP-2-L) retains only $0.126$ R@1. Crucially,
architectures designed explicitly for multi-scale processing
(Swin-L, FlexiViT-L, MViTv2-L) degrade along the same curve as
standard ViTs. Per-backbone breakdowns and a detailed comparison
appear in \cref{app:backbone_comparison}.

\subsection{Disentangling O2I From Resolution}
\label{sec:disentangle}

A plausible explanation for the curves in \cref{fig:retrieval_vs_o2i}
is resolution loss: at larger camera distances, fewer pixels cover the
object. In our controlled renders, however, distance changes both object
resolution and object-to-image ratio (O2I). We disentangle these factors
by comparing raw cross-distance retrieval to retrieval after O2I matching.
For each backbone, we evaluate all 20 ordered cross-distance pairs in two
conditions: $\alpha$, the raw R@1 where query and gallery differ in both
O2I and resolution, and $\beta$, the R@1 after tight-cropping both images
to the object bounding box and bilinearly upsampling each crop to the
backbone's native input size, so that the object fills the frame in both
images. Both conditions are evaluated against a single-distance gallery,
the 3{,}021 objects rendered at the target distance $d_t$: tight crops of
the same object at different distances collapse to near-identical images,
so a multi-distance gallery is not meaningful under $\beta$, and
evaluating $\alpha$ on the same gallery keeps the two conditions
commensurable. Absolute R@1 is therefore higher than in
\cref{fig:retrieval_vs_o2i}, which uses the 15{,}101-item multi-distance
gallery; the decomposition compares $\alpha$ and $\beta$ under one fixed
protocol. Because viewpoint, lighting, background, and centering are fixed
by construction, cropping makes the O2I of query and gallery identical, while
the resolution gap between them persists in full: the far crop is
magnified from roughly $3\times$ fewer captured pixels than the near
one and remains commensurately blurrier, since upsampling adds no
information. Within $\beta$ the two images of a pair therefore differ
only in resolution, and the failure that survives is attributable to
resolution alone. We define the remaining failure after O2I matching,
$\delta=1-\beta$, as the resolution-only residual in this controlled
setting, and the raw failure, $\gamma=1-\alpha$, as the combined effect of
O2I mismatch and resolution loss; the resolution share
$\varepsilon=\delta/\gamma$ is the fraction of cross-distance failure
explained by resolution alone. Patch-level pooling probes that
complement this analysis appear in \cref{app:selective_pool}.

\begin{figure}[t]
  \centering
  \includegraphics[width=\linewidth]{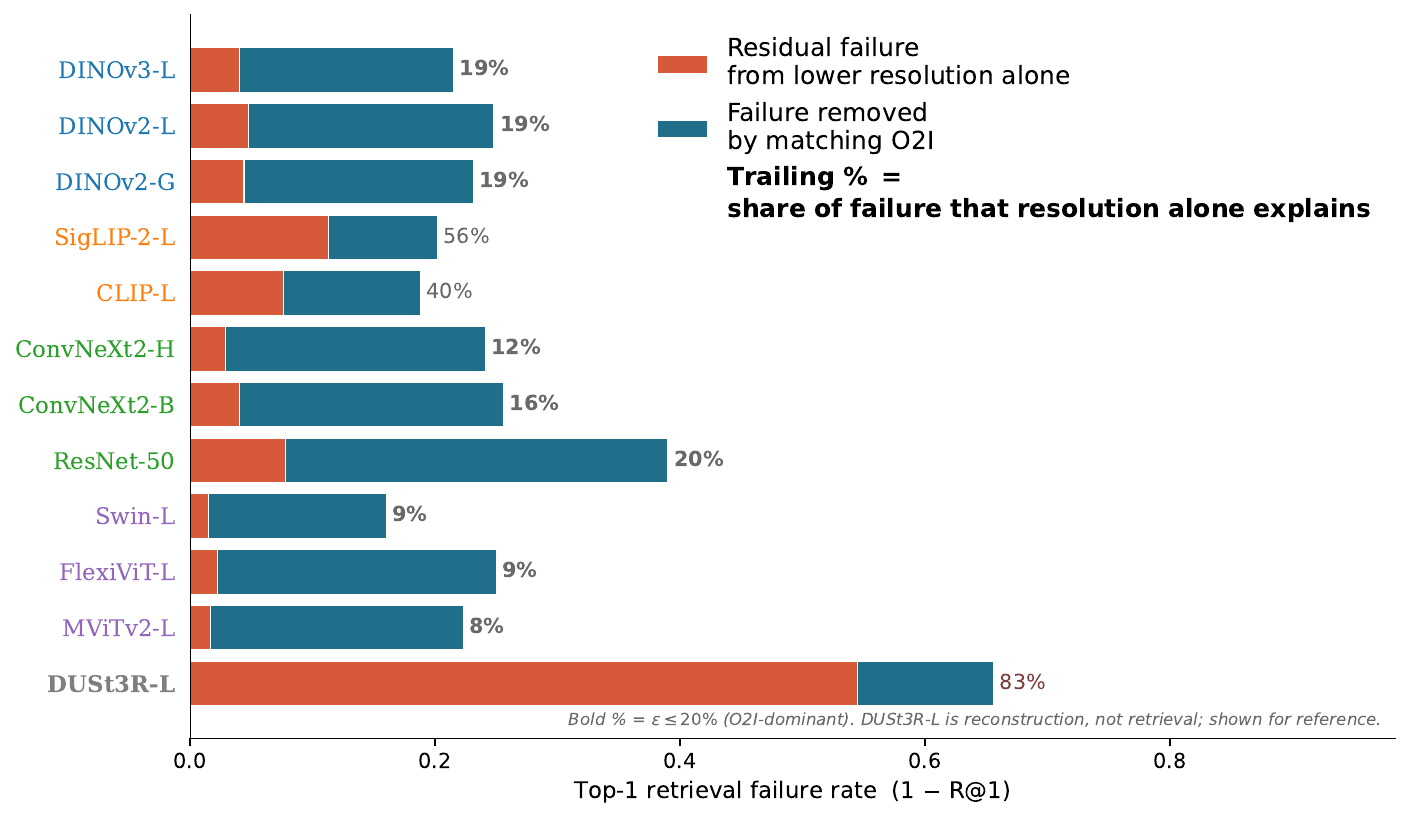}
  \caption{\textbf{Disentangling O2I from resolution across backbones.}
  X-axis: cross-distance Top-1 failure rate ($1-\text{R@1}$) on
  3{,}021 Objaverse objects. We tight-crop both query and gallery
  to the object bbox to match O2I. The \textbf{teal} segment is the
  failure recovered by this matching. The \textbf{orange} segment
  remains, attributable to resolution alone. Trailing \% is orange's
  share of the total. For 9 of 12 backbones tested it is
  less than $20\%$: O2I mismatch dominates. Multi-scale backbones
  have small resolution shares yet remain equally susceptible to O2I.
  Per-backbone numbers: \cref{tab:disentangle}
  (\cref{app:disentangle_table}).}
  \label{fig:disentangle_chart}
\end{figure}

\cref{fig:disentangle_chart} reports the decomposition across 11
retrieval backbones plus DUSt3R-L (reconstruction, included for
reference) on 3{,}021 Objaverse objects, each evaluated at its
native input resolution; per-backbone numbers are in
\cref{tab:disentangle} (\cref{app:disentangle_table}).
For 9 of 12 backbones tested, resolution accounts for at most
20\% of the total cross-distance degradation.
For DINOv3-L, for instance, the total drop is $0.215$ while
the resolution-only residual is only $0.040$ (19\% of the
total); the remaining 81\% vanishes once O2I is matched.
The two outliers are vision-language models
(SigLIP-2-L at 56\%, CLIP-L at 41\%), suggesting that
contrastive image-text training leaves these backbones more
sensitive to local pixel information than self-supervised ViTs
of the same size. DUSt3R-L shows the opposite pattern (83\%
resolution share), consistent with a reconstruction backbone's
reliance on pixel detail.

\textbf{Hierarchical/multi-scale architectures provide an
informative control}: designed for multi-scale processing, they cut the
resolution-only residual to 0.015--0.022 (Swin-L, FlexiViT-L,
MViTv2-L---resolution accounts for 8--9\% of their total drop).
Yet their total drop remains large (0.160--0.250), so they are
still equally susceptible to O2I mismatch. Architectural
resolution invariance does not transfer to O2I invariance,
reinforcing that the two failure modes are mechanistically
distinct. The fixed-pixel experiment of \cref{sec:canvas} complements
the decomposition: it moves O2I while the object's pixels are frozen
exactly.

\subsection{Decoupling O2I From Resolution With Fixed Pixels}
\label{sec:canvas}

The decomposition above matches O2I by cropping and upsampling,
which alters the input pixels. To move O2I while holding
resolution literally fixed, we take a tight object crop of each
object (mask bounding box at the closest render distance, resized
once so its longest side is 256\,px) and paste the identical,
unresized pixels at the center of gray canvases of growing size
(256 to 1536\,px), lowering the O2I ratio from 1 to 1/36 while
the object's pixels remain byte-identical in every condition.
DINOv3-L processes each canvas at its native size.
\cref{tab:canvas} reports the mean cosine similarity between the
features of the identical crop on different canvases. The
identical object drifts from itself as the O2I gap widens: at the
largest gap, self-similarity falls to 0.77. Nothing about the
object's pixels differs; only the fraction of the image it
occupies does. The two factors are colinear at acquisition, and
conflated under ``scale change'', but they are separate
properties of the representation.

\subsection{Cross-Distance Retrieval is Asymmetric in O2I}
\label{sec:asymmetry}

Beyond magnitude, the cross-distance failure has a directional
component.
\cref{fig:symmetry} plots, per backbone, the mean MRR with a
wide query against a tight gallery (MRR$_{q>t}$, x-axis) versus
the reverse direction (MRR$_{q<t}$, y-axis), aggregated over the
10 ordered cross-distance pairs in each direction. Every backbone
sits above the $y=x$ diagonal: a query whose object is larger
than the gallery's matches more reliably than the reverse, with
$\Delta$ ranging from $+0.020$ (SigLIP-2-L) to $+0.153$
(FlexiViT-L). The effect holds across self-supervised,
vision-language, supervised, and hierarchical/multi-scale
families---consistent with a representation-level rather than
family-specific cause. The scatter is shown in
\cref{fig:symmetry} (\cref{app:symmetry_table}); per-backbone numbers are in
\cref{tab:symmetry} (\cref{app:symmetry_table}); per-pair MRR
curves are in \cref{app:symmetry_curves}.

\section{Confirming the Diagnosis on ILIAS 100M}
\label{sec:augmentation}

If O2I mismatch is the operative cause of cross-distance
retrieval failure, an intervention that equalizes it should help
on real benchmarks. We test this on
ILIAS~100M (\cref{tab:ilias_100m}) and ROxford/RParis
(\cref{app:roxford_rparis}); the results confirm the prediction
and, as a side effect, achieve state of the art on ILIAS~100M
without training or gallery modification.

\paragraph{Setup.}
\label{sec:ilias_setup}
We evaluate on ILIAS~\citep{ilias2025} with DINOv3-L, a
top-performing model on its leaderboard: 1{,}232 queries, 4{,}715 positives,
100M YFCC distractors, official DINOv3-L/16 gallery features
(768\,px), extracted once without augmentation.

\subsection{Approach}

\paragraph{Gray-pad averaging (GP).}
For each scale $s \in \{1.0, 0.75, 0.5, 0.33, 0.25\}$, we resize
the query to $(sW, sH)$, center it on a neutral gray canvas
(RGB~$(128, 128, 128)$) of the original size, extract the
backbone's CLS feature, average
the per-scale features, and $\ell_2$-normalize, producing a
single query vector of unchanged dimension: index storage and
search cost are identical to the baseline, and the only overhead
is the extra query-side forward passes. This presents the
query at multiple apparent sizes, spanning a $4{\times}$ range.

\paragraph{Masked Scale Augmentation (MSA).}
At $s{=}0.25$ most patches in the input are gray, yet the CLS
token still attends to them. MSA uses a sparser scale set
$\{1.0, 0.5, 0.25\}$ and, at each scale, adds an additive
attention mask of $-\infty$ on every gray patch key (applied
before softmax in every transformer layer), forcing the CLS
token to attend only to object-bearing patches. Cost is
$3{\times}$ query-time FLOPs; the mask itself is negligible.
MSA relies on DINOv3's register
tokens~\citep{simeoni2025dinov3,darcet2024registers} to absorb
displaced attention.

\subsection{Results}

\cref{tab:ilias_100m} presents pre-reranking results.
GP improves R@1 from 0.550 to 0.582 ($+5.8$\% relative), and MSA improves
it further to 0.592 ($+7.7$\%); MSA also lifts R@5 from 0.625 to
0.683 ($+9.3$\%) and mAP@1000 from 0.265 to \textbf{0.292}
(\textbf{$+10.2$\%}). All improvements over the baseline are
statistically significant ($p<0.001$); paired-test $p$-values for
mAP@1000 are in \cref{tab:ilias_100m_full}
(\cref{app:ilias_100m_full}); the full testing methodology,
including McNemar tests for the recall metrics, is in
\cref{app:bootstrap}.
A simple training-free intervention motivated by the diagnostic
findings yields a $+2.7$\,pp absolute mAP improvement (10\%
relative) on the current leaderboard model, supporting the
conclusion that O2I mismatch is a practically addressable
bottleneck.
The computational overhead is $3{\times}$ query-time FLOPs, and
the gains are additive to any downstream reranking since they
raise the oracle ceiling.

\begin{table}[t]
  \centering
  \caption{\textbf{Simple query- and gallery-side object-to-image
  matching interventions achieve state of the art on ILIAS~100M.}
  DINOv3-L backbone, 1{,}232 queries, $\sim$100M gallery. Oracle =
  upper bound for any reranker on the DINOv3-L top-1000 shortlist.
  Full table with query/gallery-side cost breakdown and paired
  $t$-test $p$-values: \cref{tab:ilias_100m_full}
  (\cref{app:ilias_100m_full}).}
  \label{tab:ilias_100m}
  \small
  \begin{tabular}{lc}
    \toprule
    Method & mAP@1000 \\
    \midrule
    \multicolumn{2}{l}{\emph{Pre-reranking (first-stage retrieval)}} \\
    DINOv3-L (baseline)                          & 26.5 \\
    DINOv3-L + GP (query-side scale averaging)   & 28.3 \\
    \textbf{DINOv3-L + MSA (query-side, masked)} & \textbf{29.2} \\
    \midrule
    \multicolumn{2}{l}{\emph{Reranking over top-1000 shortlist}} \\
    Leaderboard best: PE-L (adapt.) + AMES~\citep{suma2024ames} & 39.4 \\
    \textbf{+ OWLv2 crop rerank (gallery-side)}        & \textbf{42.0} \\
    \midrule
    Oracle ceiling (perfect rerank over top-1000)      & 50.9 \\
    \bottomrule
  \end{tabular}
\end{table}

\paragraph{Cross-dataset generalization.}
MSA generalizes beyond ILIAS: on the Revisited
Oxford5K~\citep{radenovic2018revisiting} and Paris6K benchmarks,
it improves mAP on both Medium ($+0.8$\,pp Oxford, $+0.7$\,pp Paris) and
Hard ($+1.6$\,pp Oxford, $+1.3$\,pp Paris) protocols, with the largest
gains where query--gallery O2I mismatch is most severe
(\cref{app:roxford_rparis}). ILIAS remains the primary benchmark
because both its queries and gallery positives carry box
annotations, so the query--gallery O2I mismatch is directly
measurable; ROxford/RParis annotate only the queries, and
therefore serve as a reference.

\subsection{Gains Are Concentrated Where O2I Mismatch Exists}

Stratifying ILIAS queries by their query-vs-positive O2I ratio
(median $3.25{\times}$; 93\% of queries are tighter than their
positives), \cref{tab:scale_conditioned} shows MSA gains
concentrate where they should: the reverse-direction bin
($<\!1{\times}$) is slightly degraded, moderate mismatches gain
most ($+4.4$ to $+5.5$\,pp R@1), and extreme ratios
($>\!10{\times}$) gain less because the scale grid bottoms out
at $0.25{\times}$. The pattern confirms the intervention is
O2I-specific rather than generic ensembling
(\cref{app:pair_stratification}).

\paragraph{Scale is not generic test-time augmentation.}
At equal 5-view cost, GP yields $+3.2$\,pp R@1 while the best
non-scale augmentation ensemble yields only $+0.8$\,pp
(\cref{app:scale_vs_tta}); alternative scale-pooling strategies
are near-equivalent to GP and all dominated by MSA
(\cref{app:ablation}).

\paragraph{Comparison with query expansion.}
$\alpha$QE~\citep{chum2007total}, the standard query-side
enhancement, lifts mAP (26.5 to 29.1) but degrades R@5/R@10 and
cannot fix a wrong top match; our interventions improve every
metric, and the two stack to \textbf{31.4} mAP@1000
(\cref{tab:aqe}, \cref{app:aqe_table}). Mirroring the
augmentations onto the \emph{gallery} side instead hurts
retrieval, exactly as the O2I analysis predicts, while an oracle
gallery-side zoom helps strongly (\cref{app:gallery_side}).

\subsection{Learnable Alternative: LoRA Fine-Tune}
\label{sec:lora}

GP and MSA address O2I at test time. We additionally test whether
training can build O2I robustness directly into the backbone:
rank-8 LoRA adapters~\citep{hu2022lora} on every DINOv3-L
attention layer, trained to match the frozen baseline's CLS on
the canonical view while the student sees a gray-padded,
attention-masked, lower-O2I view (full recipe in
\cref{app:lora}). On the
full ILIAS 100M index, the adapted backbone reaches
R@1\,=\,60.2 and mAP@1000\,=\,29.7, matching or exceeding the
test-time interventions (MSA: 59.2/29.2) with a single forward
pass per image (training details and proxy-gallery ablations in
\cref{app:lora}). The training-time result confirms the diagnosis
from a third angle: O2I robustness can be learned rather than
imposed at query time.

\subsection{Reranking-Stage Confirmation: Object-Centric Crops}
\label{sec:rerank}

The diagnosis makes the same prediction at the reranking stage:
localizing the object in each top-$k$ candidate and re-encoding
it at a canonical scale removes the O2I mismatch by construction,
so reranking should improve (recent work reports the same
pattern with a Grounding-DINO~\citep{liu2024grounding}
region-proposal variant~\citep{choi2026patchify}).

\paragraph{Approach.}
Given the top-1000 first-stage candidates from DINOv3-L, we use
OWLv2~\citep{minderer2024owlv2} in class-agnostic objectness
mode (no text query; objectness above $0.02$, top-50 boxes per
image) to localize candidate regions in each gallery image. Each
region is cropped, padded to a square, and re-encoded by frozen
DINOv3-L, giving a query-independent per-region feature database
computed once per gallery image. The reranking score of a (query,
gallery) pair is the maximum cosine similarity between the
query's original feature and any region feature; the first-stage
score is unused. No training; gallery-side detection and crop
encoding are amortized across all queries.

\paragraph{Results.}
The reranking rows of \cref{tab:ilias_100m} report rerank metrics
on ILIAS 100M.
OWLv2 crop reranking improves mAP@1000 from 26.5 (DINOv3-L
first stage) to \textbf{42.0}, an absolute gain of
\textbf{+15.5\,pp} (+58\% relative).
R@1 improves from 55.0 to 76.0 (+21.0\,pp).
This sets a new state of the art for ILIAS 100M reranking,
exceeding the best reported leaderboard result (PE-L with linear
adaptation + AMES reranking~\citep{suma2024ames}, 39.4 mAP@1000)
by +2.6\,pp mAP.
The oracle ceiling for any top-1000 reranker on this shortlist
is 50.9 mAP, indicating substantial remaining headroom for
shortlist-aware methods.
A mechanism-level probe (\cref{app:o2i_probe}) uses an
attention-mask intervention that removes outside-bbox context
without changing any pixels; it attributes at least 51\% of the
crop-rerank gain on ILIAS and 65\% on INSTRE to O2I
normalization itself, with the rest due to outside-bbox context.

\section{Discussion and Conclusion}
\label{sec:discussion}

We analyzed O2I mismatch as an isolated failure mode of visual
instance retrieval: equalizing O2I removes most of the
cross-distance failure, the failure is asymmetric in direction,
and scale-invariant architectures do not transfer that invariance
to O2I. The training-free interventions on ILIAS~100M achieve
state of the art (29.2 pre-rerank, 42.0 post-rerank),
corroborating the analysis, and a lightweight LoRA fine-tune
(\cref{sec:lora}) matches them at single-forward-pass cost.
Scaling this to full pretraining is the natural next step toward
retrieval backbones that are O2I-invariant by construction.

%

\subsection*{Reproducibility statement}
The synthetic benchmark is fully specified in \cref{sec:benchmark}
(Objaverse object set, five camera distances, fixed isometric
viewpoint, Kaolin rasterization); backbone checkpoints and input
resolutions are listed in \cref{app:backbones}. ILIAS experiments
use the official queries, gallery, and released DINOv3-L/16
gallery features; the GP and MSA interventions are defined by
their scale sets and attention masks in \cref{sec:augmentation},
the LoRA recipe in \cref{app:lora}, and the reranker protocol,
with its objectness threshold and box cap, in \cref{sec:rerank}.
Significance testing methodology is described in
\cref{app:bootstrap}. Rendering, evaluation, and intervention
code will be released, together with the exact Objaverse object
identifiers and the category-selection procedure.

\subsection*{AI use statement}
In this work, we used generative AI tools as general-purpose aids
for drafting and polishing parts of the paper, for identifying
related work, and for writing and editing code, including the
implementation of evaluation code (implement methods). We have
reviewed all AI-assisted work and take
responsibility for the final content of this work, including text,
claims, and artifacts produced with the aid of generative AI.

\bibliographystyle{iclr2027_conference}
\bibliography{refs}

\newpage
\appendix
\section*{Supplementary Material}
\addcontentsline{toc}{section}{Supplementary Material}

\section{Synthetic Benchmark: Example Renderings}
\label{app:benchmark_example}

\begin{figure}[!ht]
  \centering
  \includegraphics[width=\linewidth]{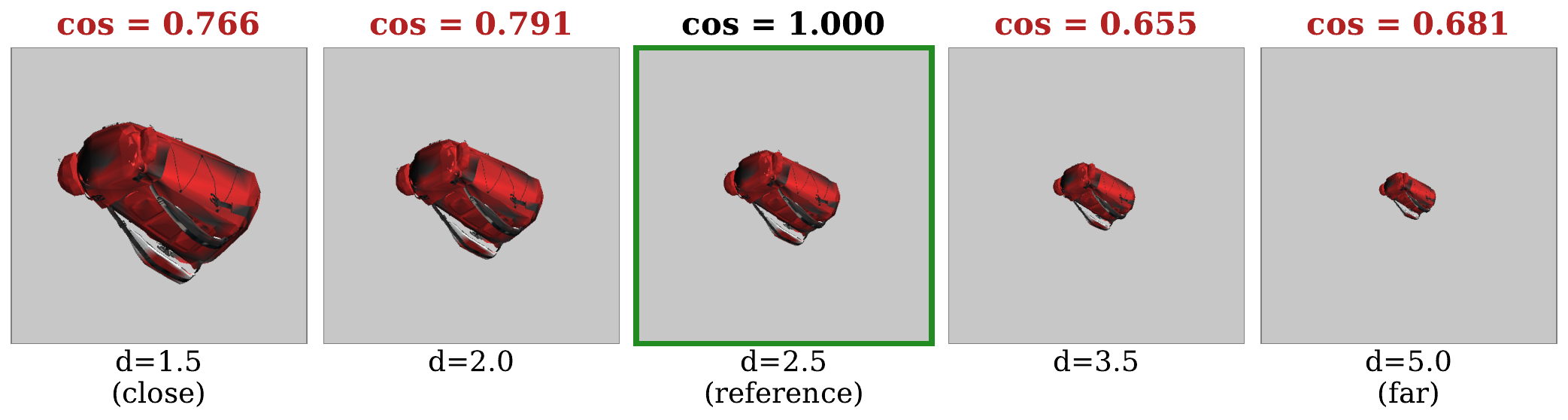}
  \caption{\textbf{Example renderings from the controlled
  benchmark.} The same backpack rendered from the same viewpoint
  at five camera distances; each label shows the DINOv3-L cosine
  similarity to the reference image (green border, $d{=}2.5$).
  Cosine drops below 0.8 at every off-reference distance, despite
  the object and viewpoint being identical---changing only the
  object's fraction of the frame is enough to break feature
  similarity.}
  \label{fig:benchmark_example}
\end{figure}

\cref{fig:benchmark_example} shows a representative object from
our controlled benchmark, rendered at the five camera distances
described in \cref{sec:benchmark} of the main paper. Across
distances the object identity, viewpoint, lighting, and
background are held fixed; only the fraction of the image the
object occupies (its O2I ratio) varies. The cosine similarity
of the per-distance feature to the reference rendering ($d{=}2.5$)
drops below 0.8 at every off-reference distance for DINOv3-L,
illustrating the cross-distance similarity decay analyzed in
\cref{sec:results}.

\section{Dataset Diversity: Categories and Instances}
\label{app:montage}

\begin{figure}[!ht]
  \centering
  \includegraphics[width=\linewidth]{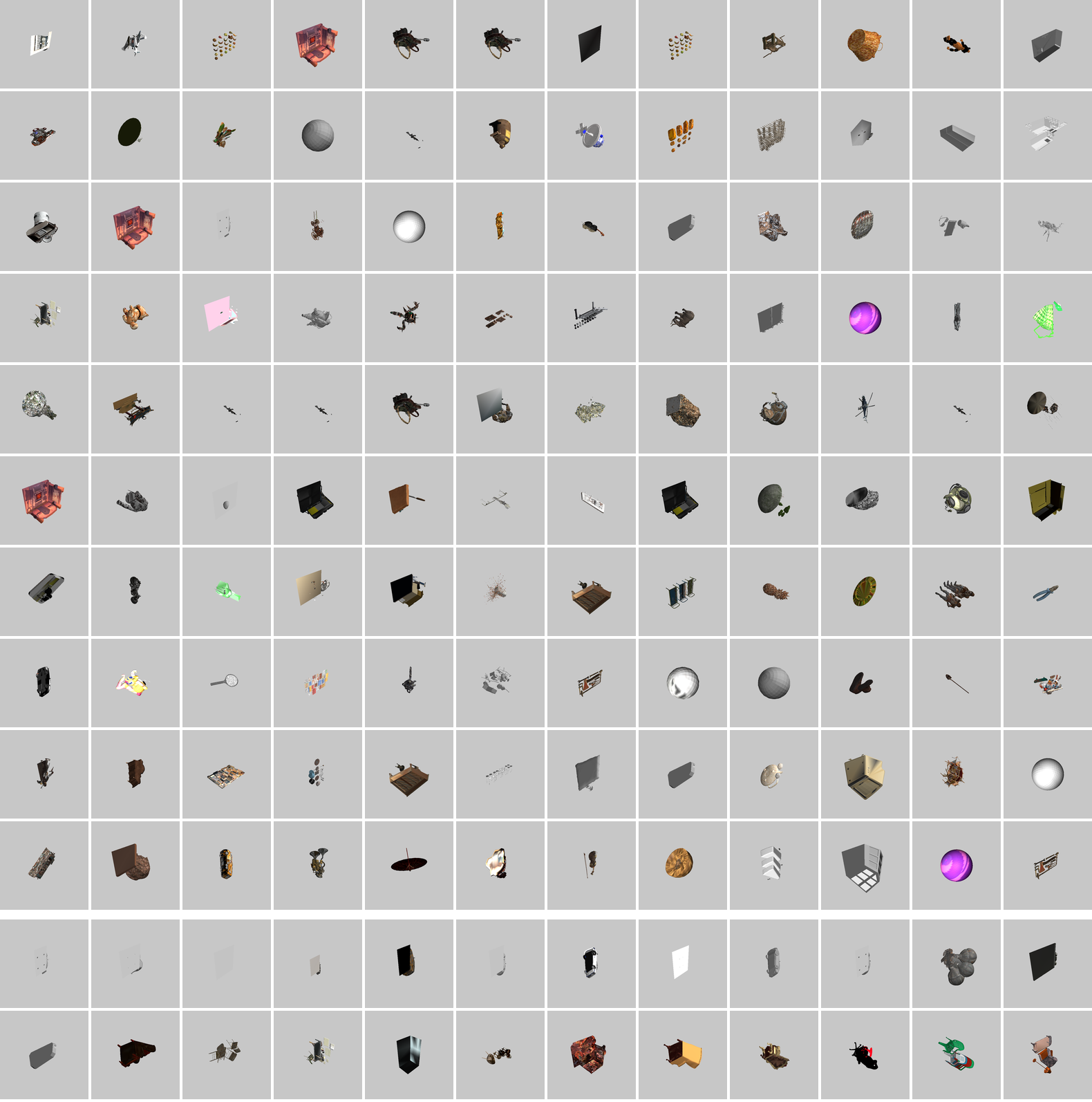}
  \caption{\textbf{Benchmark diversity.} Top block: one instance
  from each of the 120 categories. Bottom two rows: instances of
  two example categories, illustrating within-category variety.
  All images are the reference-distance ($d{=}2.5$) renders.}
  \label{fig:montage}
\end{figure}

\cref{fig:montage} shows the diversity of the benchmark: one
instance from each of the 120 categories (top block) and multiple
instances of two example categories (bottom rows), all rendered at
the reference distance $d{=}2.5$. The benchmark contains 3{,}021
objects across 120 categories, ${\sim}25$ instances per category.

\section{Disentangle Decomposition: Per-Backbone Numbers}
\label{app:disentangle_table}

\cref{fig:disentangle_chart} of the main paper visualizes the
disentangle decomposition. \cref{tab:disentangle} below reports the
full per-backbone numerical breakdown.

\begin{table}[t]
  \centering
  \caption{\textbf{Disentangling O2I from resolution across backbones} (3,021 Objaverse objects, mean cross-distance R@1 over 20 ordered distance pairs; both conditions evaluated against single-distance galleries of 3{,}021 items as described in \cref{sec:disentangle}, so absolute values are higher than the multi-distance-gallery numbers of \cref{tab:sr_metrics}). $\alpha$: standard retrieval (query and gallery differ in both object-to-image fraction, O2I, and pixel resolution). $\beta$: same retrieval after both query and gallery are tight-cropped to the object bbox and bilinearly upsampled to the backbone's native input size, so only resolution differs. Derived columns ($\gamma$, $\delta$, $\varepsilon$) are computed per the formulas in the header; the O2I share is $1-\varepsilon$. For 9 of the 12 evaluated backbones (9 of 11 retrieval backbones), resolution accounts for $\leq$20\% of the cross-distance failure. Hierarchical/multi-scale models (Swin-L, FlexiViT-L, MViTv2-L) have the smallest resolution-only residual, yet their O2I-only drops are comparable to standard ViTs of similar size---architectural scale invariance does not transfer to O2I invariance. DUSt3R-L is included as a reference (reconstruction, not retrieval); its failure is dominated by resolution loss, the opposite of the dominant retrieval pattern.}
  \label{tab:disentangle}
  \small
  \setlength{\tabcolsep}{4pt}
  \begin{tabular}{@{}lcccccc@{}}
    \toprule
    Backbone & \shortstack{Input\\px} & \shortstack{$\alpha$\\Raw cross-dist.\\R@1} & \shortstack{$\beta$\\R@1 after\\O2I matching} & \shortstack{$\gamma=1-\alpha$\\Total\\drop} & \shortstack{$\delta=1-\beta$\\Resolution-only\\residual} & \shortstack{$\varepsilon=\delta/\gamma$\\Resolution\\share} \\
    \midrule
    \multicolumn{7}{l}{\emph{Self-supervised}} \\
    \quad DINOv3-L & 224 & 0.785 & 0.960 & 0.215 & 0.040 & 19\% \\
    \quad DINOv2-L & 224 & 0.752 & 0.952 & 0.248 & 0.048 & 19\% \\
    \quad DINOv2-G & 224 & 0.769 & 0.956 & 0.231 & 0.044 & 19\% \\
    \multicolumn{7}{l}{\emph{Vision-language}} \\
    \quad SigLIP-2-L & 384 & 0.798 & 0.887 & 0.202 & 0.113 & 56\% \\
    \quad CLIP-L & 336 & 0.812 & 0.924 & 0.188 & 0.076 & 40\% \\
    \multicolumn{7}{l}{\emph{Supervised}} \\
    \quad ConvNeXt2-H & 512 & 0.759 & 0.971 & 0.241 & 0.029 & 12\% \\
    \quad ConvNeXt2-B & 224 & 0.744 & 0.960 & 0.256 & 0.040 & 16\% \\
    \quad ResNet-50 & 224 & 0.610 & 0.922 & 0.390 & 0.078 & 20\% \\
    \multicolumn{7}{l}{\emph{Hierarchical/multi-scale}} \\
    \quad Swin-L & 384 & 0.840 & 0.985 & 0.160 & 0.015 & 9\% \\
    \quad FlexiViT-L & 240 & 0.750 & 0.978 & 0.250 & 0.022 & 9\% \\
    \quad MViTv2-L & 224 & 0.777 & 0.983 & 0.223 & 0.017 & 8\% \\
    \multicolumn{7}{l}{\emph{Reconstruction}} \\
    \quad DUSt3R-L & 512 & 0.344 & 0.455 & 0.656 & 0.545 & 83\% \\
    \bottomrule
  \end{tabular}
\end{table}

\section{Fixed-Pixel Canvas Experiment: Cosine Matrix}
\label{app:canvas_table}

\cref{tab:canvas} reports the full cosine-similarity matrix for the
fixed-pixel canvas experiment of \cref{sec:canvas} in the main paper.

\begin{table}[!ht]
  \centering
  \caption{\textbf{O2I moves, resolution frozen.} Mean cosine
  similarity between DINOv3-L features of the \emph{identical}
  object crop pasted on gray canvases of different sizes
  (3{,}021 objects; bootstrap 95\% CIs within $\pm0.003$). Only
  the fraction of the image occupied by the object changes across
  canvases; the object's pixels are byte-identical.}
  \label{tab:canvas}
  \small
  \begin{tabular}{lcccc}
    \toprule
    Cosine & c@256 & c@512 & c@1024 & c@1536 \\
    \midrule
    c@256  & 1.00 & 0.96 & 0.87 & 0.77 \\
    c@512  & 0.96 & 1.00 & 0.94 & 0.83 \\
    c@1024 & 0.87 & 0.94 & 1.00 & 0.94 \\
    c@1536 & 0.77 & 0.83 & 0.94 & 1.00 \\
    \bottomrule
  \end{tabular}
\end{table}

\section{Per-Backbone Asymmetry Numbers}
\label{app:symmetry_table}

\begin{figure}[!ht]
  \centering
  \includegraphics[width=0.7\linewidth]{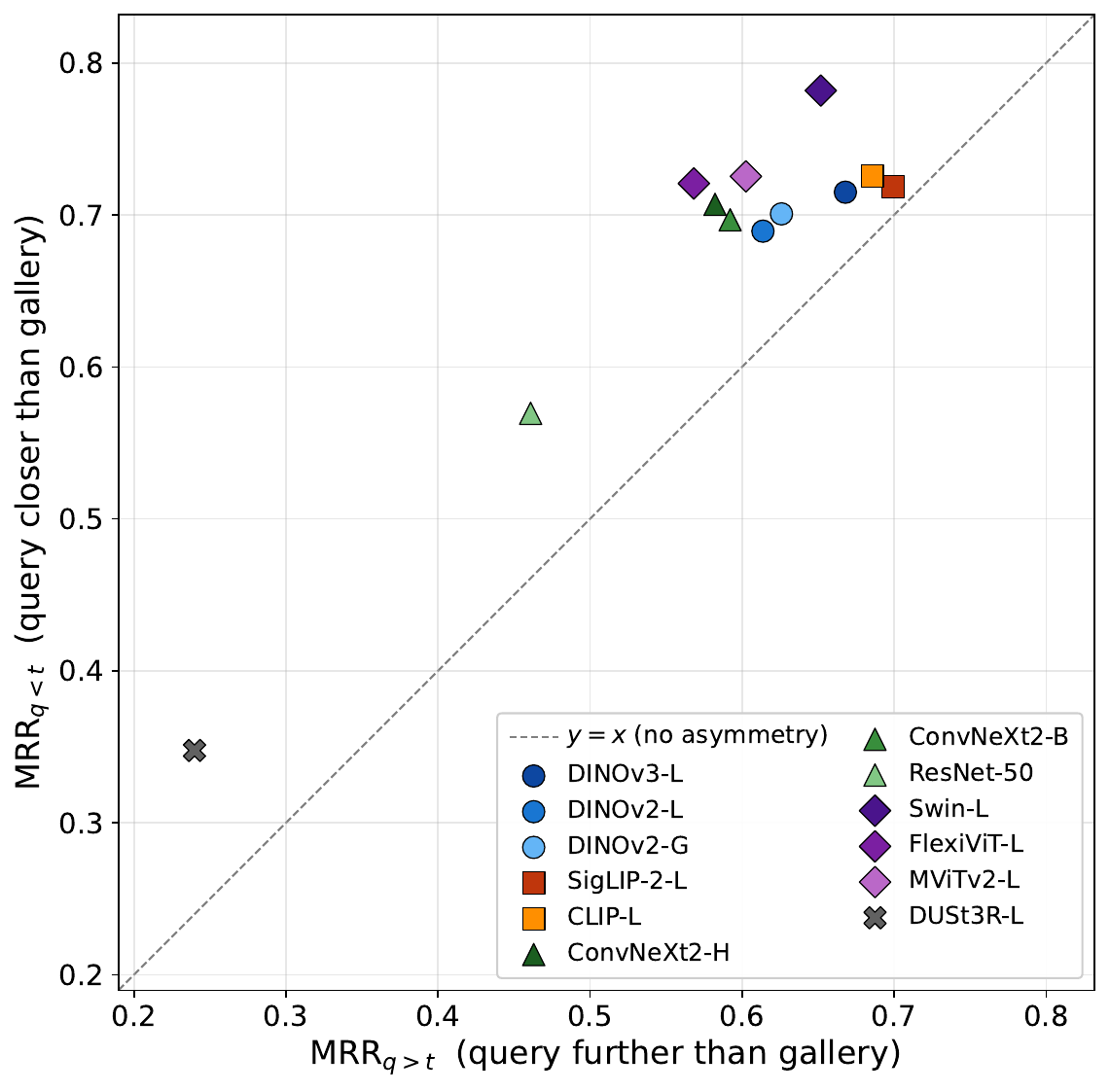}
  \caption{\textbf{Cross-distance retrieval is asymmetric in O2I.}
  Each axis reports mean MRR over the 10 ordered cross-distance
  pairs on our synthetic Objaverse benchmark. The x-axis is the
  direction in which the query is farther from the object than the
  gallery (MRR$_{q>t}$); the y-axis is the reverse direction
  (MRR$_{q<t}$, query closer than the gallery). The dashed
  diagonal is $y=x$ (no asymmetry). Each marker is one backbone;
  marker shape encodes the family, color shade the specific
  backbone within the family. Every backbone sits above the
  diagonal: a tight query against a wide gallery matches more
  reliably than the reverse.}
  \label{fig:symmetry}
\end{figure}

\cref{tab:symmetry} reports the per-backbone mean MRR underlying
\cref{fig:symmetry} of the main paper.

\begin{table}[t]
  \centering
  \caption{\textbf{Cross-distance retrieval is asymmetric in O2I.}
  Mean MRR per backbone over the ten ordered $(d_q, d_t)$
  cross-distance pairs in each direction. MRR$_{q<t}$ (``tight
  query''): query closer than target, query O2I $>$ target O2I.
  MRR$_{q>t}$ (``wide query''): query farther than target, query
  O2I $<$ target O2I. $\Delta$ = MRR$_{q<t}$ $-$ MRR$_{q>t}$ is
  positive for every backbone: a tight query against a wide
  gallery matches more reliably than the reverse.}
  \label{tab:symmetry}
  \small
\begin{tabular}{llccc}
\toprule
Family & Backbone & MRR$_{q<t}$ & MRR$_{q>t}$ & $\Delta$ \\
 & & \scriptsize tight query & \scriptsize wide query & \\
\midrule
Self-supervised & DINOv3-L & 0.715 & 0.668 & \textbf{+0.047} \\
 & DINOv2-L & 0.689 & 0.614 & \textbf{+0.075} \\
 & DINOv2-G & 0.701 & 0.626 & \textbf{+0.075} \\
\midrule
Vision-language & SigLIP-2-L & 0.719 & 0.699 & \textbf{+0.020} \\
 & CLIP-L & 0.726 & 0.686 & \textbf{+0.040} \\
\midrule
Supervised & ConvNeXt2-H & 0.707 & 0.582 & \textbf{+0.125} \\
 & ConvNeXt2-B & 0.697 & 0.592 & \textbf{+0.105} \\
 & ResNet-50 & 0.569 & 0.461 & \textbf{+0.108} \\
\midrule
Hierarchical & Swin-L & 0.782 & 0.652 & \textbf{+0.130} \\
 & FlexiViT-L & 0.721 & 0.568 & \textbf{+0.153} \\
 & MViTv2-L & 0.725 & 0.602 & \textbf{+0.123} \\
\midrule
Reconstruction & DUSt3R-L & 0.348 & 0.240 & \textbf{+0.108} \\
\bottomrule
\end{tabular}

\end{table}

\section{Per-Backbone Symmetry Curves}
\label{app:symmetry_curves}

\begin{figure}[t]
  \centering
  \includegraphics[width=\linewidth]{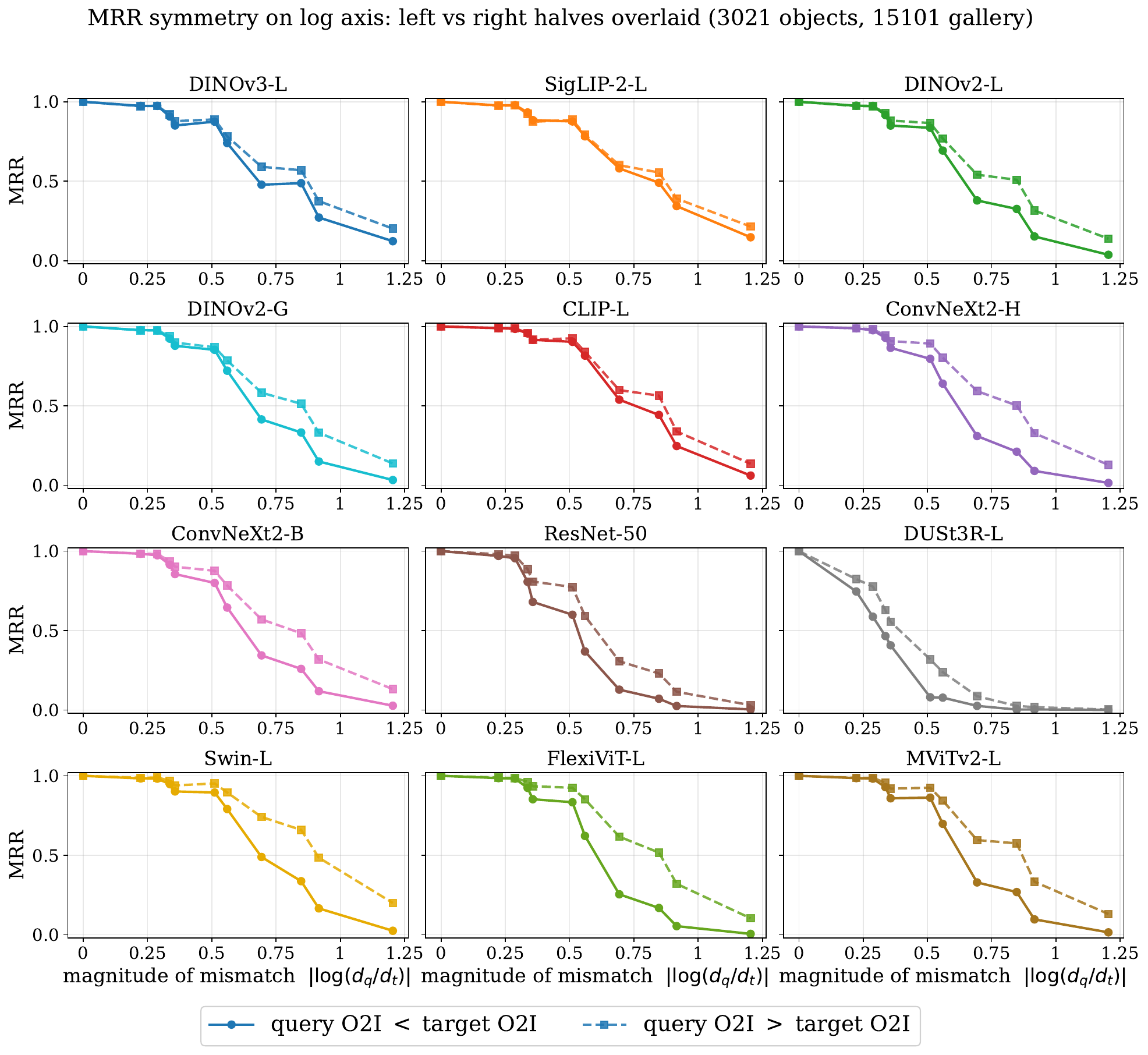}
  \caption{\textbf{Cross-distance retrieval asymmetry, per-pair
  MRR curves.} Each panel: one backbone. Y-axis: MRR. X-axis:
  $|\log(d_q/d_t)|$. Dashed: tight query ($d_q<d_t$); solid: wide
  query ($d_q>d_t$).}
  \label{fig:symmetry_panels}
\end{figure}

\cref{tab:symmetry} in the main paper summarises the
cross-distance asymmetry as a single mean-MRR per direction per
backbone. \cref{fig:symmetry_panels} shows the same data
disaggregated by mismatch magnitude: each panel overlays both
directions of mismatch for one backbone, with MRR on the y-axis
and the magnitude of mismatch $|\log(d_q/d_t)|$ on the x-axis;
\textbf{dashed} curves are query O2I~$>$~target O2I (tight query,
$d_q<d_t$) and \textbf{solid} curves are the reverse, so pairs at
the same $x$ have identical mismatch magnitude. The dashed curve
sits clearly above the solid one across the entire range for
every retrieval backbone, confirming that the asymmetry is not an
artefact of averaging over magnitudes.

\section{Additional Backbone-Comparison Results}
\label{app:backbone_comparison}

\cref{tab:sr_metrics,tab:sr_compare} report the per-backbone
breakdown underlying the summary discussion in
\cref{sec:results} of the main paper. No single
factor---architecture, model size, or training
objective---fully predicts O2I robustness, but several patterns
emerge:
\begin{itemize}
  \item \textbf{Hierarchical/multi-scale models.}
    Swin-L achieves the highest overall mean R@1 (0.677) and
    leads at moderate gaps ($|\Delta d|{=}1.0{-}2.5$), showing
    that hierarchical processing provides a genuine advantage.
    However, FlexiViT-L (0.604) and MViTv2-L (0.623) are no
    better than standard ViTs, and all three collapse at
    extreme gaps ($|\Delta d|{=}3.5$: Swin 0.073, MViTv2 0.044,
    FlexiViT 0.030). Multi-scale architecture mitigates but
    does not solve O2I mismatch.
  \item \textbf{Vision-language models} (CLIP-L, SigLIP-2-L) are
    the most robust at extreme gaps. CLIP-L leads at small gaps;
    SigLIP-2-L is superior at $|\Delta d| \geq 2.5$.
    We hypothesize that web-scale image-text training, where
    the same concept appears at widely varying apparent sizes,
    encourages scale-tolerant representations---though
    differences in training data, objectives, and input
    resolution make this difficult to isolate cleanly.
  \item \textbf{Model size is not decisive.}
    DINOv3-L outperforms DINOv2-G (1.1B~params) despite being
    3.6$\times$ smaller; ConvNeXt2-H (657M) and ConvNeXt2-B
    (128M) perform nearly identically.
  \item \textbf{Reconstruction features} (DUSt3R-L) collapse to
    near-zero R@1 at large gaps ($\le 0.030$ for $|\Delta d| \geq 2.0$),
    confirming that dense stereo features are unsuitable for retrieval
    across varying O2I ratios.
\end{itemize}

\begin{table}[t]
  \centering
  \caption{\textbf{O2I robustness: overall retrieval metrics.} Mean over all 20 cross-distance pairs (3,021 objects, 15,101 gallery items). Best per column in \textbf{bold}.}
  \label{tab:sr_metrics}
  \small
  \begin{tabular}{@{}lcccc@{}}
    \toprule
    \textbf{Model} & \textbf{R@1}$\uparrow$ & \textbf{R@5}$\uparrow$ & \textbf{R@10}$\uparrow$ & \textbf{MRR}$\uparrow$ \\
    \midrule
    \multicolumn{5}{l}{\emph{Self-supervised}} \\
    \quad DINOv3-L & 0.645 & 0.744 & 0.784 & 0.691 \\
    \quad DINOv2-L & 0.608 & 0.700 & 0.740 & 0.652 \\
    \quad DINOv2-G & 0.620 & 0.712 & 0.753 & 0.663 \\
    \multicolumn{5}{l}{\emph{Vision-language}} \\
    \quad SigLIP-2-L & 0.662 & 0.763 & \textbf{0.804} & 0.709 \\
    \quad CLIP-L & 0.664 & 0.752 & 0.790 & 0.706 \\
    \multicolumn{5}{l}{\emph{Supervised}} \\
    \quad ConvNeXt2-H & 0.603 & 0.691 & 0.729 & 0.645 \\
    \quad ConvNeXt2-B & 0.600 & 0.693 & 0.734 & 0.644 \\
    \quad ResNet-50 & 0.466 & 0.567 & 0.611 & 0.515 \\
    \multicolumn{5}{l}{\emph{Hierarchical/multi-scale}} \\
    \quad Swin-L & \textbf{0.677} & \textbf{0.764} & 0.799 & \textbf{0.717} \\
    \quad FlexiViT-L & 0.604 & 0.689 & 0.726 & 0.644 \\
    \quad MViTv2-L & 0.623 & 0.710 & 0.748 & 0.664 \\
    \multicolumn{5}{l}{\emph{Reconstruction}} \\
    \quad DUSt3R-L & 0.256 & 0.332 & 0.366 & 0.294 \\
    \bottomrule
  \end{tabular}
\end{table}

\begin{table}[t]
  \centering
  \caption{\textbf{O2I robustness: R@1 by camera distance gap.} Cross-distance retrieval over 3,021 objects at 5 camera distances. $|\Delta d|$ is the absolute gap between query and target camera distances. The O2I mismatch ratio $\mathrm{O2I}_q/\mathrm{O2I}_t = (d_t/d_q)^2$ is \emph{not} monotone in $|\Delta d|$ on this discrete grid (e.g., $|\Delta d|{=}2.5$ corresponds to a $4{\times}$ ratio, smaller than the $5.44{\times}$ at $|\Delta d|{=}2.0$); accordingly, R@1 at $|\Delta d|{=}2.5$ exceeds R@1 at $|\Delta d|{=}2.0$ for most backbones, consistent with the paper's claim that O2I drives the failure. Gallery: target at $d_{\mathrm{target}}$ + all other objects at all distances (15,101 items). Best per column in \textbf{bold}.}
  \label{tab:sr_compare}
  \small
  \resizebox{\linewidth}{!}{%
  \begin{tabular}{@{}lcccccccc@{}}
    \toprule
    \textbf{Model} & $|\Delta d|$=0.5 & $|\Delta d|$=1.0 & $|\Delta d|$=1.5 & $|\Delta d|$=2.0 & $|\Delta d|$=2.5 & $|\Delta d|$=3.0 & $|\Delta d|$=3.5 & \textbf{Mean} \\
    \midrule
    \multicolumn{9}{l}{\emph{Self-supervised}} \\
    \quad DINOv3-L & 0.962 & 0.866 & 0.764 & \textbf{0.454} & 0.456 & 0.250 & 0.110 & 0.645 \\
    \quad DINOv2-L & 0.962 & 0.854 & 0.749 & 0.343 & 0.377 & 0.172 & 0.053 & 0.608 \\
    \quad DINOv2-G & 0.966 & 0.866 & 0.773 & 0.346 & 0.415 & 0.173 & 0.052 & 0.620 \\
    \multicolumn{9}{l}{\emph{Vision-language}} \\
    \quad SigLIP-2-L & 0.967 & 0.874 & 0.785 & 0.442 & 0.510 & \textbf{0.288} & \textbf{0.126} & 0.662 \\
    \quad CLIP-L & \textbf{0.982} & 0.913 & 0.833 & 0.419 & 0.488 & 0.218 & 0.059 & 0.664 \\
    \multicolumn{9}{l}{\emph{Supervised}} \\
    \quad ConvNeXt2-H & 0.975 & 0.857 & 0.752 & 0.287 & 0.372 & 0.153 & 0.044 & 0.603 \\
    \quad ConvNeXt2-B & 0.972 & 0.846 & 0.743 & 0.297 & 0.375 & 0.155 & 0.048 & 0.600 \\
    \quad ResNet-50 & 0.954 & 0.702 & 0.528 & 0.096 & 0.146 & 0.037 & 0.006 & 0.466 \\
    \multicolumn{9}{l}{\emph{Hierarchical/multi-scale}} \\
    \quad Swin-L & 0.979 & \textbf{0.921} & \textbf{0.844} & 0.418 & \textbf{0.536} & 0.253 & 0.073 & \textbf{0.677} \\
    \quad FlexiViT-L & 0.979 & 0.882 & 0.763 & 0.269 & 0.355 & 0.131 & 0.030 & 0.604 \\
    \quad MViTv2-L & 0.978 & 0.891 & 0.781 & 0.348 & 0.380 & 0.154 & 0.044 & 0.623 \\
    \multicolumn{9}{l}{\emph{Reconstruction}} \\
    \quad DUSt3R-L & 0.686 & 0.316 & 0.256 & 0.005 & 0.030 & 0.004 & 0.000 & 0.256 \\
    \bottomrule
  \end{tabular}}
\end{table}

\section{Cross-Dataset Validation: ROxford5K and RParis6K}
\label{app:roxford_rparis}

To verify that the gains are not specific to ILIAS, we evaluate
on the Revisited Oxford~\citep{radenovic2018revisiting} and Paris
benchmarks using the same DINOv3-L model and the standard
revisited protocol.
Queries in these benchmarks are bounding-box crops of
architectural details (median area 27\% for Oxford, 41\% for
Paris), while gallery images show full building views---a natural
O2I mismatch.

\begin{table}[t]
  \centering
  \caption{\textbf{ROxford5K and RParis6K} (DINOv3-L, 768\,px).
  MSA improves mAP across both benchmarks, with the largest gains
  on the Hard protocol where O2I mismatch is most severe.}
  \label{tab:roxford_rparis}
  \small
  \begin{tabular}{lccc|ccc}
    \toprule
    & \multicolumn{3}{c|}{ROxford5K} & \multicolumn{3}{c}{RParis6K} \\
    Method & Easy & Med & Hard & Easy & Med & Hard \\
    \midrule
    Original & \textbf{93.5} & 81.5 & 64.2 & 95.3 & 91.6 & 82.2 \\
    GP (5 scales) & 91.1 & 81.1 & 64.8 & 95.3 & 91.9 & 83.0 \\
    MSA & 93.1 & \textbf{82.3} & \textbf{65.8} & \textbf{95.6} & \textbf{92.3} & \textbf{83.5} \\
    \bottomrule
  \end{tabular}
\end{table}

MSA improves Medium mAP on both benchmarks (+0.8 on Oxford,
+0.7 on Paris) and provides the largest gains on the Hard
protocol (+1.6 on Oxford, +1.3 on Paris), where queries
depict small architectural details that must be matched against
full-scene gallery images.
Plain GP slightly hurts on Oxford Easy ($-2.4$\,pp) because
easy queries already have moderate O2I overlap; attention
masking in MSA largely mitigates this ($-0.4$\,pp).

\section{Query Expansion Comparison: Full Table}
\label{app:aqe_table}

\cref{tab:aqe} reports the full comparison between query-side O2I
matching and $\alpha$QE on ILIAS 100M. $\alpha$QE (best setting,
$k{=}1$) improves mAP but degrades R@5 and R@10 and does not
improve R@1: it reorders results that were already retrieved, and
cannot recover a query whose top match is wrong. GP and MSA act
on the query representation before retrieval and improve every
metric. The two are orthogonal and stack: MSA + $\alpha$QE
reaches 31.4 mAP@1000, the best first-stage result.

\begin{table}[!ht]
  \centering
  \caption{\textbf{Query-side O2I matching vs.\ query expansion}
  on ILIAS 100M (DINOv3-L). $\alpha$QE lifts mAP but cannot fix a
  wrong top match; O2I matching improves every metric, and the
  two stack.}
  \label{tab:aqe}
  \small
  \begin{tabular}{lcccc}
    \toprule
    Method & R@1 & R@5 & R@10 & mAP@1000 \\
    \midrule
    Original (baseline) & 55.0 & 62.5 & 66.8 & 26.5 \\
    GP-5   & 57.9 & 66.6 & 69.6 & 28.3 \\
    MSA-3  & 59.2 & \textbf{68.3} & \textbf{70.5} & 29.2 \\
    Original + $\alpha$QE & 55.0 & 55.8 & 57.1 & 29.1 \\
    GP-5 + $\alpha$QE     & 57.9 & 59.4 & 60.6 & 30.7 \\
    MSA-3 + $\alpha$QE    & \textbf{59.2} & 60.8 & 62.1 & \textbf{31.4} \\
    \bottomrule
  \end{tabular}
\end{table}

\section{Gallery-Side Augmentation}
\label{app:gallery_side}

The main-paper interventions act on the query only. We test the
same augmentations applied to the gallery side, on an 856K proxy
gallery (the union of the vanilla DINOv3-L top-1000 shortlists
across all 1{,}232 queries). Applying GP-5 or MSA-3 to every
gallery image \emph{lowers} R@1 by 12--13\,pp and mAP by about
7\,pp: gallery objects are already smaller than the query
objects, so shrinking them further pushes the O2I ratio in the
wrong direction. Zooming instead toward the object in each
gallery image---using its ground-truth box for positives (an
oracle unavailable in a realistic retrieval setting) and a center
crop for distractors---improves retrieval substantially: R@1
rises from 0.55 to 0.75 and mAP from 0.26 to 0.45. Both
directions confirm the diagnosis: retrieval improves exactly when
the query--gallery O2I gap shrinks, and degrades when it grows.
Because gallery-side augmentation must touch every gallery image,
the practical gallery-side intervention remains detector-guided
reranking on the shortlist (\cref{sec:rerank} of the main
paper).

\section{Alternative Query-Side Strategies}
\label{app:alt_aug}

\subsection{Scale augmentation vs.\ generic test-time augmentation}
\label{app:scale_vs_tta}

Multi-view averaging is a well-known technique that can improve
any retrieval system. To verify that scale augmentation
provides a benefit beyond generic ensembling, we compare GP
(5~scale views) against non-scale augmentations of equal
computational cost (5~views each) on the full ILIAS 100M
benchmark (\cref{tab:q4_random_aug}).
Horizontal flip, color jitter, random crops, and a mixed
non-scale augmentation all provide modest improvements
(0.3--1.0\,pp R@1), but GP provides \textbf{3.2\,pp}---roughly
three times the gain of the best non-scale alternative at the
same cost. This confirms that the improvement is predominantly
O2I-specific rather than a generic ensembling effect.

\begin{table}[t]
  \centering
  \caption{\textbf{Scale vs.\ generic augmentation} on ILIAS 100M
  (DINOv3-L). GP provides $3{\times}$ the R@1 gain of non-scale
  augmentations at equal query-time cost (5~views).}
  \label{tab:q4_random_aug}
  \small
  \begin{tabular}{lccccc}
    \toprule
    Method & Views & R@1 & R@5 & mAP \\
    \midrule
    Original & 1 & 55.0 & 62.5 & 26.5 \\
    \midrule
    Horizontal flip & 2 & 55.9 & 63.2 & 26.9 \\
    Color jitter & 5 & 55.3 & 63.0 & 26.7 \\
    Random crops & 5 & 55.5 & 62.7 & 26.4 \\
    Mixed non-scale & 5 & 55.8 & 63.5 & 27.0 \\
    \midrule
    \textbf{GP (scale)} & 5 & \textbf{58.2} & \textbf{65.9} & \textbf{28.3} \\
    \bottomrule
  \end{tabular}
\end{table}

\subsection{Ablation: alternative query-side strategies}
\label{app:ablation}

\cref{tab:ablation} compares MSA against alternative query-side
interventions on the full ILIAS 100M benchmark (DINOv3-L,
768\,px). All methods modify only the query representation; the
gallery is unchanged.

\begin{table}[t]
  \centering
  \caption{\textbf{Ablation of query-side strategies} on ILIAS 100M
  (DINOv3-L, 768\,px gallery). All methods are training-free and
  query-only. MSA achieves the best results on all metrics.}
  \label{tab:ablation}
  \small
  \begin{tabular}{lcccc}
    \toprule
    Method & R@1 & R@5 & R@10 & mAP \\
    \midrule
    Original (baseline) & 55.0 & 62.5 & 66.8 & 26.5 \\
    \midrule
    \multicolumn{5}{l}{\emph{Scale averaging variants}} \\
    \quad GP avg (5 scales) & 58.2 & 65.9 & 69.1 & 28.3 \\
    \quad Median across scales & 57.4 & 65.4 & 68.8 & 28.2 \\
    \quad Consistency-weighted mean & 58.2 & 65.9 & 69.1 & 28.3 \\
    \midrule
    \multicolumn{5}{l}{\emph{Object localization (requires bbox)}} \\
    \quad Crop$\to$fill + GP avg & 58.0 & 66.1 & 68.9 & 28.3 \\
    \midrule
    \multicolumn{5}{l}{\emph{Multi-resolution (no gray padding)}} \\
    \quad Multi-res \{768, 518, 336, 224\} & 55.7 & 63.8 & 67.9 & 27.0 \\
    \midrule
    \multicolumn{5}{l}{\emph{Attention masking (DINOv3 only)}} \\
    \quad \textbf{MSA} (3 scales + mask) & \textbf{59.2} & \textbf{68.3} & \textbf{70.5} & \textbf{29.2} \\
    \bottomrule
  \end{tabular}
\end{table}

Simple averaging (GP) provides a strong gain (+3.2\,pp R@1);
no alternative pooling strategy (median, consistency weighting)
significantly improves upon it.
Cropping to the object bounding box and filling the frame
performs comparably to GP but requires a bounding box.
Multi-resolution inference (each scale obtained by downsampling
then upsampling back to 768\,px, so every forward pass uses the
same input grid) yields a small improvement over the baseline
($+0.7$\,pp R@1, $+0.5$\,pp mAP) but is clearly weaker than
GP/MSA: augmenting resolution loss without changing O2I is
much less effective.
MSA achieves the best results on all metrics by preventing
gray patches from contaminating the representation, rather
than attempting to remove their effect post-hoc.

\section{Selective Patch Pooling and Attention-Mask Analysis}
\label{app:o2i_probe}

This appendix contains two mechanism-level probes that support
the analysis in the main paper. \cref{app:selective_pool}
examines which spatial patches contribute to the global
descriptor on the synthetic benchmark;
\cref{app:attention_mask_probe} decomposes the OWLv2 crop-rerank
gain (\cref{sec:rerank}) into an O2I component and an
outside-bbox-context component using a pixel-preserving
attention-mask intervention.

\subsection{Selective patch pooling on the synthetic benchmark}
\label{app:selective_pool}

Pooling only the patch tokens within the object bounding box
(identified by background detection on the synthetic renders)
yields R@1~=~0.793, compared to 0.845 for the CLS token and
0.617 for all-patch mean pooling.
Notably, background-only patches still achieve 0.583---far above
chance---and this \emph{increases} to 0.620 when boundary-adjacent
patches are excluded (2-patch erosion), ruling out silhouette
leakage and confirming that background tokens carry substantial
object information via long-range cross-attention.
The CLS token outperforms both, indicating that its learned
attention already performs effective (though imperfect)
selective pooling.
These results suggest that O2I sensitivity is \emph{coupled}
with the model's object encoding: changing the
object-to-background ratio does not merely add noise but shifts
the operating point of the attention-mediated representation,
though the precise mechanism likely involves multiple interacting
factors beyond token-ratio effects alone.

\paragraph{Attention masking during gray-pad.}
Blocking attention to known-uninformative gray patches during
gray-pad augmentation (described in \cref{sec:augmentation})
improves mean R@1 from 0.920 to 0.931 on the synthetic benchmark,
with the largest gain on the hardest asymmetric pair
($d{=}5.0 \to d{=}1.5$: $+14.5$\,pp).
This suggests that even when background tokens are uninformative,
the model allocates attention to them, and preventing this
recovers discriminative power.

\subsection{Decomposing OWLv2 crop-rerank gains: O2I vs outside-bbox context}
\label{app:attention_mask_probe}

OWLv2 crop reranking (\cref{sec:rerank}) conflates two distinct
changes to the gallery representation: (i)~the CLS token no
longer attends to any patches outside the object bounding box,
removing all outside-bbox context---distractor objects,
background statistics, scene layout, illumination cues; and
(ii)~the object is resized to fill the frame, resetting the
object-to-image ratio to $\sim$100\%.
To attribute the gain we isolate these factors with a
mechanism-level probe that does not alter any pixels.

\paragraph{Attention-mask probe.}
For each gallery image we pre-compute which DINOv3-L patches
overlap the ground-truth object bbox, assemble an additive
attention mask that sets $-\infty$ on every outside-bbox key,
and register a forward pre-hook that injects this mask into every
transformer layer.
The CLS token therefore cannot integrate information from any
outside-bbox patch, yet the image, patch positions, O2I
ratio, and pixel content remain untouched.
We compare this to two anchors: the full-image baseline
(everything attended, O2I unchanged) and a GT-bbox crop oracle
(outside-bbox removed, O2I~=~100\%).

\begin{table}[t]
  \centering
  \caption{\textbf{O2I vs outside-bbox context ablation} on
  INSTRE and ILIAS core (4{,}715-positive gallery, no 100M distractors;
  DINOv3-L 768\,px, GP queries).
  BBox attention masking removes outside-bbox information
  without changing the O2I ratio.
  The residual gap between the masked and cropped rows isolates
  the O2I contribution.}
  \label{tab:o2i_probe}
  \small
  \begin{tabular}{lcccc}
    \toprule
    & & & \multicolumn{2}{c}{mAP} \\
    \cmidrule(l){4-5}
    Condition & O2I & Outside-bbox & INSTRE & ILIAS \\
    \midrule
    Full image (baseline)   & unchanged   & attended & 0.854 & 0.628 \\
    BBox attention mask     & unchanged   & \textbf{masked} & 0.886 & 0.747 \\
    GT-bbox crop (oracle)   & \textbf{100\%} & removed & 0.944 & 0.873 \\
    \midrule
    Total gain              & \multicolumn{2}{c}{} & +9.0\,pp & +24.5\,pp \\
    \quad from outside-bbox & \multicolumn{2}{c}{(mask $-$ full)} & +3.2\,pp & +11.9\,pp \\
    \quad from O2I          & \multicolumn{2}{c}{(crop $-$ mask)} & +5.8\,pp & +12.6\,pp \\
    \quad O2I share         & \multicolumn{2}{c}{}  & \textbf{65\%} & \textbf{51\%} \\
    \bottomrule
  \end{tabular}
\end{table}

\paragraph{Results.}
\cref{tab:o2i_probe} decomposes the crop-oracle gain on both datasets.
On INSTRE, blocking attention to outside-bbox tokens recovers +3.2
mAP out of the +9.0 mAP total---35\% of the gap---while the
remaining \textbf{65\% requires changing the O2I ratio} by actually
resizing the object to fill the frame.
On ILIAS, with its substantially larger query-gallery O2I
mismatch (median 42\,pp), the absolute gain is +24.5\,mAP and the
split is close to even (49\% outside-bbox, \textbf{51\% O2I}).
Outside-bbox information contributes more on ILIAS because its
galleries contain more scene clutter and photometric variation;
O2I mismatch, however, remains the dominant single factor on both
datasets.
Critically, the attention-mask probe changes no pixels and
keeps patch positions fixed, so its effect cannot be attributed to
pixel-masking artefacts or to a shifted spatial prior---it
directly isolates the information the CLS token aggregates
through attention.

\paragraph{Implication.}
The fact that a pure attention-space intervention recovers at
most one-half of the crop-oracle gain establishes a non-trivial
lower bound on the contribution of O2I normalization to
object-centric retrieval.
This supports the paper's central claim: mitigating O2I
mismatch---either by reducing the query's effective O2I
(GP/MSA on the query side) or by raising the gallery's effective
O2I (object-centric crop encoding)---is the
principal driver of the observed gains, and outside-bbox
context, while helpful, is a secondary effect.

\section{O2I-Robust LoRA Fine-Tuning}
\label{app:lora}

The query-side interventions in the main paper are training-free.
We additionally test whether a lightweight, O2I-aware
fine-tune of the backbone itself can build O2I invariance into
the representation, removing the need for query-time gray-pad
averaging.

\paragraph{Approach.}
We attach LoRA adapters~\citep{hu2022lora} of rank~8 to the
\texttt{q\_proj} and \texttt{v\_proj} matrices in every
self-attention layer of DINOv3-L, leaving all other parameters
frozen (\texttt{lora\_alpha=16}, \texttt{lora\_dropout=0.1}).
Training uses a self-distillation objective: the frozen DINOv3-L
processes the original image at $512\times512$ and produces a
target CLS token; the LoRA-adapted student processes the same
image after gray-pad rescaling at a random gray-pad fraction
$s \sim \mathcal{U}(0.1, 0.3)$ (which lowers the object's O2I
ratio in the input) and is trained with an InfoNCE objective in
which positive pairs are (student, teacher) for the same image
and negatives are the other images in the batch, using cosine
similarity scaled by temperature $\tau = 0.07$. During the student
forward pass, attention to the gray-padded patches is masked at
every layer (the same mechanism as the MSA query-side
intervention); the teacher receives the original image
unmasked. This amounts to telling the model: ``a lower-O2I view
of the same object, with attention restricted to the visible
content, should produce the same global descriptor as the
canonical view.''
Training data is 25{,}000 unlabeled COCO images for a single
epoch (${\sim}55$\,minutes on one RTX A5000), with AdamW
(weight decay $0.01$), batch size 24, learning rate
$2{\times}10^{-4}$, and a ReduceLROnPlateau scheduler
(factor $0.5$, patience $3$).
Training is performed at $512$\,px to fit a larger batch on a
single RTX A5000; inference uses the ILIAS protocol of 768\,px
largest side, matching the encoding used for vanilla DINOv3-L
on the same gallery. At test time the LoRA student does a
single forward pass per image with no query-side averaging or
attention masking; the intervention lives entirely in the
adapted weights.

\paragraph{Evaluation protocol.}
The main result re-encodes the queries and the full ILIAS 100M
gallery with the adapted backbone and evaluates on the complete
index, reaching R@1~=~60.2 and mAP@1000~=~29.7 with a single
forward pass per image (\cref{sec:lora} of the main paper).
During development we additionally use an 856{,}089-image proxy
gallery for ablations:
the union, across all 1{,}232 ILIAS queries, of the top-1{,}000
nearest neighbors retrieved from the full 100M index using the
\emph{official} ILIAS DINOv3-L feature dump (4{,}715 ILIAS
positives plus 851{,}374 unique distractors). For evaluation we
re-encode this proxy gallery (and the queries) under our own
pipeline---HuggingFace \texttt{AutoModel} with the ILIAS
protocol (768\,px largest side, bf16, $\ell_2$-normalized
CLS)---and use the same encoder for the vanilla and LoRA rows,
so the within-gallery comparison is consistent. Two effects
prevent a direct equation with the full 100M numbers: (i) our
re-encoding pipeline differs slightly from the official feature
dump, which produces vanilla numbers ${\sim}0.7$\,pp below the
ILIAS leaderboard; (ii) any image our pipeline would have ranked
into the top-1{,}000 of full 100M but the official pipeline did
not is missing from the proxy. We therefore report this proxy
strictly as a within-pipeline comparison of vanilla vs.\ LoRA
vs.\ MSA/GP on a shared ${\sim}\!10^6$-scale gallery; absolute
mAP@1000 should not be compared against the leaderboard cell.

\paragraph{Proxy-gallery ablations.}
On the 856K proxy gallery the LoRA fine-tune attains
R@1~=~58.1, R@5~=~66.2 and mAP@1000~=~27.9 with a \emph{single}
forward pass per image, against vanilla DINOv3-L at
R@1~=~54.6, mAP@1000~=~25.8 on the same gallery
(\cref{tab:lora_856k})---a gain of \textbf{+3.5\,pp R@1} and
\textbf{+2.1\,pp mAP@1000}.
For reference we also score the two query-side test-time
interventions reported in the main paper on the same gallery:
GP (5-scale gray-pad averaging) reaches mAP@1000~=~27.2 and
MSA (3-scale gray-pad with attention masking) reaches
mAP@1000~=~27.9, indistinguishable from LoRA on mAP.
LoRA exceeds GP at one-fifth the query-time cost, and matches
MSA at one-third---without any test-time augmentation. This
shows that O2I robustness can be learned during training rather
than imposed at query time: the representation itself becomes
O2I-invariant. LoRA and MSA are complementary
directions---LoRA modifies what the representation \emph{is},
MSA modifies how it is \emph{queried}---and a combination is a
natural next step.

\begin{table}[h]
\centering
\small
\setlength{\tabcolsep}{4pt}
\begin{tabular}{lcccccc cc}
\toprule
 & Q-enc & G-enc & FwdPass & R@1 & R@5 & MRR & \multicolumn{2}{c}{mAP@1000} \\
\cmidrule(lr){8-9}
 & & & & & & & 856K proxy & full 100M \\
\midrule
Vanilla DINOv3-L          & vanilla & vanilla & 1 & 54.6 & 62.1 & 58.4 & 25.8 & 26.5 \\
+ GP (5 scales)           & vanilla & vanilla & 5 & 56.3 & 65.1 & 60.4 & 27.2 & 28.3 \\
+ MSA (3 scales + mask)   & vanilla & vanilla & 3 & 57.9 & 65.9 & 61.7 & 27.9 & \textbf{29.2} \\
+ LoRA (rank 8, q,v)      & LoRA    & LoRA    & 1 & \textbf{58.1} & \textbf{66.2} & \textbf{61.9} & \textbf{27.9} & \textbf{29.7} \\
\bottomrule
\end{tabular}
\caption{O2I-aware interventions on the 856{,}089-image ILIAS
top-1{,}000 proxy gallery (1{,}232 queries, 4{,}715 positives,
851{,}374 distractors). \emph{Q-enc} / \emph{G-enc} indicate the
encoder used for the query and gallery, respectively;
\emph{FwdPass} is the number of query-side forward passes per
image. GP and MSA are query-side test-time interventions against
the vanilla gallery; the LoRA row re-encodes \emph{both} query
and gallery with the LoRA-adapted weights. Replacing the LoRA
gallery with the vanilla one (cross setting) drops the LoRA row
to mAP@1000~=~24.4, confirming that the adapter shifts the
embedding space and the gallery must be re-encoded to capture
the gain. The \emph{full 100M} column reports the same backbone
runs evaluated on the complete ILIAS 100M gallery
(\cref{tab:ilias_100m}); the proxy underestimates absolute mAP
by 0.7--1.3\,pp but preserves relative ordering.}
\label{tab:lora_856k}
\end{table}

\section{ILIAS 100M Results: Full Table With Cost Breakdown and $p$-Values}
\label{app:ilias_100m_full}

\cref{tab:ilias_100m_full} reports the full version of
\cref{tab:ilias_100m} of the main paper, broken down into
query-side and gallery-side costs and including per-method
McNemar $p$-values against the DINOv3-L baseline.

\begin{table}[t]
  \centering
  \caption{\textbf{Full ILIAS 100M results.}
  DINOv3-L backbone, 1{,}232 queries, $\sim$100M gallery.
  $p$-values are from a paired $t$-test on per-query mAP@1000
  against the DINOv3-L baseline.
  ``Cost (query)'' is the online per-query cost; ``Cost (gallery)''
  is the one-time per-image cost amortized over all queries; both
  in units of one DINOv3-L forward pass at 768\,px (1{,}657 GFLOPs).
  MSA uses 3 scales with attention masking; GP uses 5 scales without
  masking. OWLv2 crop reranking uses $\sim$31 detections per image
  on average (one DINOv3-L crop encoding each, applied once per
  gallery image and cached); the query-side cost remains a single
  DINOv3-L forward.}
  \label{tab:ilias_100m_full}
  \small
  \begin{tabular}{lcrr}
    \toprule
    Method & mAP@1000 & \shortstack{Cost\\(query)} & \shortstack{Cost\\(gallery)} \\
    \midrule
    \multicolumn{4}{l}{\emph{Pre-reranking (first-stage retrieval)}} \\
    DINOv3-L (baseline)               & 26.5          & $1\times$ & $1\times$ \\
    DINOv3-L + GP (5 scales)          & 28.3          & $5\times$ & $1\times$ \\
    \quad\scriptsize$p$-value vs.\ baseline & \scriptsize $<$0.001 & & \\
    \textbf{DINOv3-L + MSA}           & \textbf{29.2} & $3\times$ & $1\times$ \\
    \quad\scriptsize$p$-value vs.\ baseline & \scriptsize $<$0.0001 & & \\
    \midrule
    \multicolumn{4}{l}{\emph{Reranking over top-1000 shortlist}} \\
    Leaderboard best: PE-L (adapt.) + AMES~\citep{suma2024ames}
                                       & 39.4          & --- & --- \\
    \textbf{+ OWLv2 crop rerank}       & \textbf{42.0} & $1\times$ & $\sim$32$\times$ \\
    \midrule
    Oracle ceiling (perfect rerank over top-1000)
                                       & 50.9          & --- & --- \\
    \bottomrule
  \end{tabular}
\end{table}

\section{Significance Testing Methodology}
\label{app:bootstrap}

Significance is assessed with McNemar's test for
binary recall metrics (R@$k$) and a paired $t$-test for mAP@1000
(computed per-query as average-precision-at-1000, then aggregated
across queries).

\section{Query-Positive Pair Stratification}
\label{app:pair_stratification}

\begin{table}[!ht]
  \centering
  \caption{\textbf{O2I-conditioned gains on ILIAS 100M.}
  MSA improvement over baseline (absolute pp), stratified by query-to-gallery
  object O2I ratio. Gains are concentrated where the query
  object is larger than the gallery's; the small reverse-direction
  bin ($<\!1{\times}$, query smaller) is slightly degraded.}
  \label{tab:scale_conditioned}
  \small
  \begin{tabular}{lrccccc}
    \toprule
    & & \multicolumn{2}{c}{Baseline} & \multicolumn{3}{c}{MSA $\Delta$} \\
    \cmidrule(lr){3-4} \cmidrule(lr){5-7}
    O2I ratio & $n$ & R@1 & mAP & $\Delta$R@1 & $\Delta$R@5 & $\Delta$mAP \\
    \midrule
    $<\!1{\times}$ (query smaller) & 85 & 74.1 & 34.6 & $-$2.4 & $-$2.4 & $-$0.7 \\
    $1$--$3{\times}$ & 479 & 62.8 & 33.7 & +4.4 & +5.4 & +2.5 \\
    $3$--$10{\times}$ & 543 & 52.3 & 22.5 & +5.5 & +7.4 & +3.6 \\
    $>\!10{\times}$ & 125 & 23.2 & 10.9 & +2.4 & +5.6 & +1.8 \\
    \midrule
    All & 1232 & 55.0 & 26.5 & +4.2 & +5.8 & +2.7 \\
    \bottomrule
  \end{tabular}
\end{table}

The O2I stratification in \cref{tab:scale_conditioned} of the main
paper buckets each \emph{query} by the ratio of the query object's
bounding-box
area to the mean bounding-box area of its positive gallery set.
This aggregates over a query's positive set (on average
$\sim\!5$ images per query on ILIAS), which smooths out
within-query variability in positive-image framing. A stricter
query-positive-pair stratification---bucketing each individual
(query, positive) pair by its own area ratio---would refine the
analysis further and is a natural extension.

\section{Backbone Implementation Details}
\label{app:backbones}

All models run in bfloat16 precision on a single NVIDIA RTX A5000.
\cref{tab:backbone_details} lists architecture, parameter count,
and pooling mode for each backbone.

\begin{table}[h]
  \centering
  \caption{\textbf{Backbone implementation details.} Pooling
  modes: \emph{CLS} = final-layer class token;
  \emph{pooled} = model's pooled vision-tower output;
  \emph{timm CLS} = pre-logit CLS embedding via timm;
  \emph{mean} = mean over all patch tokens from the encoder.}
  \label{tab:backbone_details}
  \small
  \resizebox{\linewidth}{!}{%
  \begin{tabular}{lllrl}
    \toprule
    Paradigm & Model & HF / timm name & Params & Pooling \\
    \midrule
    \multirow{3}{*}{Self-supervised}
      & DINOv2-L & \texttt{facebook/dinov2-large} & 303M & CLS \\
      & DINOv2-G & \texttt{facebook/dinov2-giant} & 1.1B & CLS \\
      & DINOv3-L & \texttt{facebook/dinov3-vitl16-pretrain-lvd1689m} & 303M & CLS \\
    \midrule
    \multirow{2}{*}{Vision-language}
      & CLIP-L (@336) & \texttt{openai/clip-vit-large-patch14-336} & 304M & pooled \\
      & SigLIP-2-L (@384) & \texttt{google/siglip2-large-patch16-384} & 316M & pooled \\
    \midrule
    \multirow{3}{*}{Supervised}
      & ConvNeXt V2-H & \texttt{facebook/convnextv2-huge-22k-512} & 657M & pooled \\
      & ConvNeXt V2-B & \texttt{facebook/convnextv2-base-22k-224} & 128M & pooled \\
      & ResNet-50 & torchvision IMAGENET1K\_V2 & 25M & avg-pool \\
    \midrule
    \multirow{3}{*}{Hierarchical/multi-scale}
      & Swin-L & \texttt{swin\_large\_patch4\_window12\_384.ms\_in22k} & 197M & timm CLS \\
      & FlexiViT-L & \texttt{hf-hub:timm/flexivit\_large.600ep\_in1k} & 304M & timm CLS \\
      & MViTv2-L & \texttt{hf-hub:timm/mvitv2\_large\_cls.fb\_inw21k} & 218M & timm CLS \\
    \midrule
    Reconstruction
      & DUSt3R-L & \texttt{naver/DUSt3R\_ViTLarge\_BaseDecoder\_512\_dpt} & 303M & mean \\
    \bottomrule
  \end{tabular}}
\end{table}

\paragraph{Notes.}
Hierarchical/multi-scale architectures were included specifically
to test whether designs aimed at scale robustness
(Swin-L: shifted-window attention over hierarchical
stages~\citep{liu2021swin}; FlexiViT-L: random patch-size training
for patch-resolution invariance~\citep{beyer2023flexivit};
MViTv2-L: progressive pooling attention~\citep{li2022mvitv2})
also confer O2I robustness.

\end{document}